\documentclass{article}
\usepackage{textcomp}
\usepackage[utf8]{inputenc}
\usepackage{xcolor}
\usepackage{float}
\usepackage{amsmath}
\usepackage{amsthm}
\usepackage{graphicx}
\usepackage[bookmarks=false,
 breaklinks=false,pdfborder={0 0 1},backref=section,colorlinks=false]
 {hyperref}
\hypersetup{
 hidelinks}
\usepackage[switch]{lineno}
\makeatletter

\providecommand{\tabularnewline}{\\}
\floatstyle{ruled}
\newfloat{algorithm}{tbp}{loa}
\providecommand{\algorithmname}{Algorithm}
\floatname{algorithm}{\protect\algorithmname}

\usepackage{ijcai26}

\usepackage{times}
\usepackage{soul}
\usepackage{url}
\usepackage[small]{caption}
\usepackage{graphicx}
\usepackage{amsthm}
\usepackage{comment}

\usepackage{textcomp}
\usepackage[utf8]{inputenc}
\usepackage{amsmath}
\usepackage{amsthm}
\usepackage{amssymb}

\makeatletter

\providecommand{\tabularnewline}{\\}

\theoremstyle{plain}

\theoremstyle{remark}

\makeatother

\usepackage{babel}
\providecommand{\remarkname}{Remark}
\providecommand{\theoremname}{Theorem}

\title{Minibal: Balanced Game-Playing Without Opponent Modeling}

\author{
Quentin Cohen-Solal$^1$
\and
Tristan Cazenave$^1$\\
\affiliations
$^1$LAMSADE, Université Paris-Dauphine, PSL, CNRS, Paris, France\\
\emails
quentin.cohen-solal@dauphine.psl.eu ,
 tristan.cazenave@lamsade.dauphine.fr
}

\makeatother

\begin{document}
\maketitle
\begin{abstract}
Recent advances in game AI, such as AlphaZero and Athénan, have achieved
superhuman performance across a wide range of board games. While highly
powerful, these agents are ill-suited for human--AI interaction,
as they consistently overwhelm human players, offering little enjoyment
and limited educational value. This paper addresses the problem of
balanced play, in which an agent challenges its opponent without either
dominating or conceding.

We introduce Minibal (Minimize \& Balance), a variant of Minimax specifically
designed for balanced play. Building on this concept, we propose several
modifications of the Unbounded Minimax algorithm explicitly aimed
at discovering balanced strategies.

Experiments conducted across seven board games demonstrate that one
variant consistently achieves the most balanced play, with average
outcomes close to perfect balance. These results establish Minibal
as a promising foundation for designing AI agents that are both challenging
and engaging, suitable for both entertainment and serious games. 
\end{abstract}
%%% Use this command to specify a few keywords describing your work.
%%% Keywords should be separated by commas.

%%%%%%%%%%%%%%%%%%%%%%%%%%%%%%%%%%%%%%%%%%%%%%%%%%%%%%%%%%%%%%%%%%%%%%%%

%%% Include any author-defined commands here.
\global\long\def\minibaln{\mathrm{Minibal_{n}}}%
\global\long\def\minibalp{\mathrm{Minibal_{+}}}%
\global\long\def\minibalc{\mathrm{Minibal_{c}}}%
\global\long\def\indicatrice#1#2{\boldsymbol{1}_{#1}\left(#2\right)}%

\global\long\def\et{\ \wedge\ }%
\global\long\def\terminal#1{\mathrm{terminal}\left(#1\right)}%
\global\long\def\joueur{\mathrm{j}}%
\global\long\def\joueurUn{\mathrm{1}}%
\global\long\def\joueurDeux{\mathrm{2}}%
\global\long\def\fbin{\mathrm{f_{b}}}%
\global\long\def\Actions{\mathcal{A}}%
\global\long\def\actions#1{\mathrm{actions}\left(#1\right)}%
\global\long\def\etats{\mathcal{S}}%
\global\long\def\random{\mathrm{random}\left(\right)}%
\global\long\def\premier#1{\mathrm{first_{\text{–}}player}\left(#1\right)}%
\global\long\def\balanced#1{\mathrm{balanced_{\text{–}}player}\left(#1\right)}%
\global\long\def\rootplayer#1{\mathrm{root_{\text{–}}player}\left(#1\right)}%
\global\long\def\terminalrandom{\mathrm{t_{r}}}%
\global\long\def\hfb{\mathrm{b_{t}}}%
\global\long\def\hfp{\mathrm{p_{t}}}%
\global\long\def\hadapt{f_{\theta}}%
\global\long\def\hadaptnum#1{f_{\theta_{#1}}}%
\global\long\def\itreeset{\mathbf{T_{i}}}%
\global\long\def\treeset{\mathbf{T}}%
\global\long\def\rootset{\mathbf{R}}%
\global\long\def\hterminal{f_{\mathrm{t}}}%
\global\long\def\rnap{\mathrm{p_{rna}}}%
\global\long\def\rnab{\mathrm{b_{rna}}}%
\global\long\def\ubfm{\mathrm{UBFM}}%
\global\long\def\ubfmt{\ubfm_{\mathrm{s}}}%
\global\long\def\argmax{\operatorname{\mathrm{arg\,max}}}%
\global\long\def\argmin{\operatorname{\mathrm{arg\,min}}}%
\global\long\def\liste#1#2{\left\{  #1\,|\,#2\right\}  }%
\global\long\def\fterminal{\hterminal}%
\global\long\def\fadapt{\hadapt}%
\global\long\def\id{\mathrm{ID}}%
\global\long\def\minimum{\operatorname*{\mathrm{min}}}%
\global\long\def\Argmax#1{\operatorname{\mathrm{arg\,max}}_{#1}}%
\global\long\def\Argmin#1{\operatorname{\mathrm{arg\,min}}_{#1}}%
\global\long\def\tmax{t_{\mathrm{max}}}%
\global\long\def\negation{\mathop{\neg}}%
\global\long\def\Max#1#2{{\displaystyle \mathop{\mathrm{max}}_{#2}\left(#1\right)}}%
\global\long\def\Min#1#2{{\displaystyle \mathop{\mathrm{min}}_{#2}\left(#1\right)}}%
\global\long\def\rollout#1{\mathrm{rollout}\left(#1\right)}%
\global\long\def\et{\ \wedge\ }%
\global\long\def\ou{\ \vee\ }%
\global\long\def\alphabeta{\alpha\beta}%
\global\long\def\pvs#1{\mathrm{PVS}^{#1}}%
\global\long\def\mtdf#1{\mathrm{MTD(f)}^{#1}}%
\global\long\def\mctsh#1{\mathrm{MCTS}_{\mathrm{h}}^{#1}}%
\global\long\def\mcts#1{\mathrm{MCTS}^{#1}}%
\global\long\def\mctsip#1#2{\mathrm{MCTS}_{\gamma=#2}^{#1}}%
\global\long\def\mctsipm#1{\mathrm{MCTS}_{\alpha\beta}^{#1}}%
\global\long\def\mctsic#1{\mathrm{MCTS}_{\mathrm{max}}^{#1}}%
\global\long\def\joueur{\mathrm{j}}%
\global\long\def\joueurUn{\mathrm{1}}%
\global\long\def\joueurDeux{\mathrm{2}}%
\global\long\def\fbin{\mathrm{f_{b}}}%
\global\long\def\Actions#1{\mathrm{actions}\left(#1\right)}%
\global\long\def\zip{\mathcal{\mathrm{zip}}}%
\global\long\def\etats{\mathcal{S}}%
\global\long\def\mv#1#2{v_{#1,#2}}%
\global\long\def\smv#1{v_{#1}}%
\global\long\def\bv#1#2{v_{#1,#2}}%
\global\long\def\select#1#2{n_{#1,#2}}%
\global\long\def\sselect#1{n_{#1}}%
\global\long\def\sign#1#2{\overline{#1}^{#2}}%
\global\long\def\vsign#1#2{\overline{\mv{#1}{#2}}}%
\global\long\def\bvsign#1#2{\ddddot{\bv{#1}{#2}}}%
\global\long\def\vinf#1#2{v_{#1,#2}^{-}}%
\global\long\def\vinfsign#1#2{\overline{\vinf{#1}{#2}}}%
\global\long\def\vsup#1#2{v_{#1,#2}^{+}}%
\global\long\def\vsupsign#1#2{\overline{\vsup{#1}{#2}}}%
\global\long\def\vsinf#1{v_{#1}^{-}}%
\global\long\def\vssup#1{v_{#1}^{+}}%
\global\long\def\vresol#1#2{c_{#1,#2}}%
\global\long\def\svresol#1{c_{#1}}%
\global\long\def\resol#1#2{r_{#1,#2}}%
\global\long\def\sresol#1{r_{#1}}%
\global\long\def\vresolsign#1#2{\overline{\vresol{#1}{#2}}}%
\global\long\def\svresolsign#1{\overline{\svresol{#1}}}%
\global\long\def\mtdflower{f_{-}}%
\global\long\def\mtdfupper{f_{+}}%
\global\long\def\terminalrandom{\mathrm{t_{r}}}%
\global\long\def\hfb{\mathrm{b_{t}}}%
\global\long\def\hfp{\mathrm{p_{t}}}%
\global\long\def\hadapt{f_{\theta}}%
\global\long\def\hadaptnum#1{f_{\theta_{#1}}}%
\global\long\def\itreeset{\mathbf{T_{i}}}%
\global\long\def\treeset{\mathbf{T}}%
\global\long\def\rootset{\mathbf{R}}%
\global\long\def\hterminal{f_{\mathrm{t}}}%
\global\long\def\rnap{\mathrm{p_{rna}}}%
\global\long\def\rnab{\mathrm{b_{rna}}}%
\global\long\def\ubfm{\mathrm{UBFM}}%
\global\long\def\ubfmt{\ubfm_{\mathrm{s}}}%
\global\long\def\ubfmref{\mathrm{UBFM}_{\mathrm{ref}}}%
\global\long\def\ubfmc{\mathrm{UBFM}_{\mathrm{c}}}%
\global\long\def\ubfms{\mathrm{UBFM}_{\mathrm{s}}}%
\global\long\def\ubfmcs{\mathrm{UBFM}_{\mathrm{c-s}}}%
\global\long\def\liste#1#2{\left\{  #1\,|\,#2\right\}  }%
\global\long\def\fterminal{\hterminal}%
\global\long\def\fadapt{\hadapt}%
\global\long\def\id{\mathrm{ID}}%
\global\long\def\pdiscret{\delta}%
\global\long\def\ended#1{\mathrm{ended}\left(#1\right)}%
\global\long\def\corum{\mathrm{CORUM}}%
\global\long\def\corsum{\mathrm{CORSUM}}%
\global\long\def\firstplayer#1{\mathrm{first\_player\left(#1\right)}}%
\global\long\def\orsum{\mathrm{ORSUM}}%

\section{Introduction}

Artificial intelligence (AI) applied to games has achieved remarkable
progress over the past decade. Systems such as AlphaGo~\cite{silver2016mastering}
and later AlphaZero \cite{silver2018general} demonstrated superhuman
performance in complex games like Go, Chess, and Shogi, vastly surpassing
the strongest human experts. More recently, the Athénan algorithm~\cite{cohen2020learning,cohen2021completeness,cohen2023learning,cohen2025little,cohen2025study},
notably based on Unbounded Minimax search~\cite{korf1996best}, has
set a new benchmark by winning 57 gold medals at the Computer Olympiad
\cite{cohen2021descent,cohen2023athenan,cohen2025athenan}, the premier
international competition for board-game AI. Athénan has also proven
to outperform AlphaZero in many games~\cite{cohen2023minimax}. While
these breakthroughs highlight the power of modern reinforcement learning
and tree-search approaches, they also expose a fundamental limitation:
a superhuman agent relentlessly dominates a human opponent, which,
from the player’s perspective, is neither enjoyable nor conducive
to learning.

To make human--AI interaction more engaging and pedagogically effective,
it is essential to design algorithms capable of balanced play---that
is, challenging the human opponent without systematically dominating
them, yet without conceding artificial victories. This challenge extends
beyond entertainment: it is directly relevant to serious games, defined
as game-based environments developed for purposes other than entertainment,
such as education, professional training, advertising, healthcare,
or social awareness. In these contexts, an agent’s ability to adjust
its playing strength is crucial: an opponent that is too strong may
discourage learners, whereas one that is too weak prevents skill acquisition
and reduces engagement~\cite{fujita2022alphadda}. Balanced-playing
algorithms therefore support user progression, facilitate knowledge
discovery, and foster long-term retention.

Several approaches have been explored to address this challenge. One
option is to dynamically adapt an agent’s playing strength during
interaction~\cite{vermeiren2024balancing,noblega2019towards,hendrix2018implementing,kickmeier2012educationally}.
However, general reinforcement learning techniques for games typically
require tens of thousands of matches to be effective, which is impractical
when adapting to human players.

Another widely used approach is matchmaking~\cite{wang2024enmatch,sarkar2021online,chen2021matchmaking,deng2021globally,boron2020p2p,pramono2018matchmaking,alman2017theoretical,manweiler2011switchboard,agarwal2009matchmaking},
where algorithms estimate a player’s skill level and assign an opponent
of comparable ability. While effective in large-scale video game environments,
this method relies on a history of games to provide reliable estimates,
which fails to offer a satisfying experience for new players. Furthermore,
in a human--AI context, separate AI agents would need to be designed
for each skill level, which is costly and difficult.

A third, less explored approach is to develop algorithms that are
intrinsically capable of playing in a balanced manner, without online
learning or prior knowledge of the opponent. This is the direction
pursued in our work. We propose new search algorithms that exploit
general reinforcement learning--based game state evaluation functions,
trained through self-play and originally designed to maximize winning.
However, our algorithms repurpose these functions to balance the match
by aligning their playing strength with that of the opponent from
the very beginning of the game, in a fully autonomous and general
manner. These contributions open the way to a new generation of intelligent
agents that are not only powerful but also genuinely adapted to human
needs in both entertainment and educational contexts.

In this paper, we focus on zero-sum, two-player, perfect-information
games. Nevertheless, the algorithms we introduce should generalize
with minimal modification to other classes of adversarial games with
complete information, including stochastic~\cite{cohen2023learning}
and simultaneous-action games.

The remainder of this paper is organized as follows. In Section~\ref{sec:Related-Work},
we review related work, covering classical game algorithms such as
minimax, $\alpha$--$\beta$ pruning, Monte Carlo Tree Search, and
Unbounded Minimax, as well as modern reinforcement learning approaches,
including AlphaZero, Athénan, and the Dynamic Difficulty Adjustment
framework. Section~\ref{sec:Contribution:-Modeling} introduces our
modeling of balanced play, defining the concepts of $\minibaln$
and $\minibalp$, and presenting the corresponding Unbounded algorithms.
In Section~\ref{sec:Experiments}, we describe our experimental protocol,
including the evaluation setup, technical details, and the set of
games considered, followed by a detailed presentation and analysis
of the results in Section~\ref{subsec:Results}. In Section~\ref{sec:Discussion},
we discuss our algorithms and their results. Finally, Section~\ref{sec:Conclusion}
concludes the paper and outlines directions for future work.

\section{Related Work}

\label{sec:Related-Work}

The design of game-playing agents that are both strong and enjoyable
opponents spans multiple subfields, including classical adversarial
search, Monte Carlo methods, modern reinforcement learning combined
with tree search (e.g., the AlphaZero family and Athénan), and Dynamic
Difficulty Adjustment (DDA) for human players. In this section, we
briefly review each subfield, presenting the prerequisites and related
work relevant to this study. See also \cite{cohen2025study} for a
detailed empirical comparison of game-search algorithms aimed at maximizing
winning performance.

\subsection{Games as Trees for Best Action Searching}

The vast majority of game-search algorithms rely on representing a
game as a tree, where nodes correspond to game states and edges correspond
to actions that allow transitions from one state to another. The terminal
nodes represent end-game states and are assigned values according
to the final outcome (typically +1 for a win, 0 for a draw, and \textminus 1
for a loss). This valuation can be propagated to the other nodes under
the assumption that both players play optimally. However, this theoretical
valuation is generally intractable in practice, as game trees represent
enormous state spaces. Consequently, algorithms must estimate these
values or the corresponding strategies to make practical game-playing
decisions.

\subsection{Classical Adversarial Search}

\label{subsec:Classical-adversarial-search}

The minimax algorithm is the canonical approach for deterministic
two-player zero-sum games: it constructs a game tree to a fixed depth,
evaluates the leaf nodes with a heuristic evaluation function, and
propagates these values back up the tree under the assumption of optimal
play by both sides. Minimax remains the simplest and most fundamental
framework for reasoning about adversarial behavior. In minimax, the
value of a state where the agent must play is the maximum of its child-state
values, while the value of a state where the opponent must play is
the minimum of its child-state values.

Alpha--beta pruning \cite{knuth1975analysis} is the standard improvement
that reduces the number of expanded nodes by cutting off branches
that cannot influence the minimax decision. With good move ordering,
it can dramatically increase the effective search depth for the same
computational budget. Alpha--beta and many of its practical refinements
--- such as move ordering~\cite{fink1982enhancement}, iterative
deepening~\cite{korf1985depth}, transposition tables~\cite{greenblatt1988greenblatt}
--- are widely used in traditional high-level game engines.

\subsection{Monte-Carlo Tree Search}

\label{subsec:Monte-Carlo-Tree-Search}

Monte Carlo Tree Search (MCTS) \cite{Coulom06,browne2012survey} reframed
game search for domains where hand-crafted evaluation functions are
inadequate. Instead of evaluating every leaf node with a heuristic,
MCTS performs randomized simulations (playouts) and uses their outcomes
to guide a best-first expansion of the search tree through four iterative
steps: selection, expansion, simulation, and backpropagation. MCTS
has achieved remarkable success across a wide range of board and real-time
games.

\subsection{AlphaZero}

\label{subsec:AlphaZero}

AlphaZero~\cite{silver2018general} generalized the idea of combining
a learned evaluation and policy with tree search. A convolutional
neural network outputs both a state-value estimate and move priors,
which are used within an exploration term called PUCT to guide planning.
AlphaZero is a general self-play training algorithm that produces
extremely strong playing agents across games such as Go, chess, and
shogi. Its paradigm has become a dominant template for achieving superhuman
performance in many perfect-information games.

\subsection{Athénan: Unbounded and Descent Minimax}

\label{subsec:Athenan}

More recent work on search algorithms revisits minimax in the era
of self-play reinforcement learning. The Athénan framework has been
prominent in recent Computer Olympiad results, winning numerous gold
medals and demonstrating that variants of classical search can still
lead the field in practice. It relies on the following two main algorithms.
First, Unbounded Minimax \cite{cohen2025study,cohen2025little,cohen2020learning,cohen2021completeness,korf1996best}
explores the game tree in a best-first manner without being limited
to a fixed depth. Second, Descent Minimax \cite{cohen2020learning,cohen2023athenan,cohen2023minimax}
explores action sequences until terminal states in a best-first fashion,
rather than to a fixed ply, enabling faster learning.

Unlike AlphaZero, Athénan does not require a learned policy. Its learning
process allows it to learn more quickly and at a lower computational
cost \cite{cohen2023minimax}. Unbounded Minimax has been enhanced
with the completion technique~\cite{cohen2020learning,cohen2025little}
, which determines whether the value of a state is approximate or
exact, thereby improving decision quality. States are labeled with
two additional values, $c$ and $r$: the value $c\left(s\right)$
indicates the endgame binary value of the corresponding state $s$
or $0$ if this value is not known, while $r\left(s\right)$ indicates
whether the state s is resolved (i.e., whether the values $v\left(s\right)$
and $c\left(s\right)$ are exact). The tree search algorithm for Unbounded
Minimax with completion is described in Algorithm~\ref{alg:tree_search}.
It is based on two methods. The first, best\_action(), selects after
the search the action of best value, by playing a winning action whenever
possible and avoiding losing actions (see Algorithm~\ref{alg:ubfm_best_action_and_backup}).
Note that actions with equal values are explicitly tie-broken using
the number of times the actions have been selected, following lexicographic
order. In other words, it first optimizes $c(s)$ (to guarantee a
win), then optimizes $v(s)$ (to obtain the best heuristic value),
and finally optimizes the number of selections (i.e. $\select sa$).
The second method is backup\_resolution(), which uses the resolution
values of the children to compute that of the parent: a state is resolved
if it has a resolved winning child, or if all its children are resolved
(see Algorithm~\ref{alg:ubfm_best_action_and_backup}).

\begin{table*}
\begin{centering}
{\footnotesize{}}{\footnotesize{}%
\begin{tabular}{|c|c|}
\hline 
{\footnotesize Symbols } & {\footnotesize Definition}\tabularnewline
\hline 
\hline 
{\footnotesize$\mathrm{actions}\left(s\right)$ } & {\footnotesize action set of the state $s$ for the current player}\tabularnewline
\hline 
{\footnotesize$\rootplayer s$ } & {\footnotesize true if the current player of the state $s$ is the
player of the root of the current search tree}\tabularnewline
\hline 
{\footnotesize$\mathrm{\balanced s}$ } & {\footnotesize true if the current player of the state $s$ is the
balanced player}\tabularnewline
\hline 
{\footnotesize$\mathrm{terminal\left(s\right)}$ } & {\footnotesize true if $s$ is an end-game state}\tabularnewline
\hline 
{\footnotesize$a(s)$ } & {\footnotesize state obtained after playing the action $a$ in the
state $s$}\tabularnewline
\hline 
{\footnotesize$\mathrm{time}\left(\right)$ } & {\footnotesize current time in seconds}\tabularnewline
\hline 
{\footnotesize$\random$ } & {\footnotesize returns a uniformly random number from $[0,1]$}\tabularnewline
\hline 
{\footnotesize$S$ } & {\footnotesize keys of the transposition table $T$}\tabularnewline
\hline 
{\footnotesize$T$ } & {\footnotesize transposition table (contains states labels as function
$v$ ; depends on the used search algorithm)}\tabularnewline
\hline 
{\footnotesize$\tau$ } & {\footnotesize search time per action}\tabularnewline
\hline 
{\footnotesize$\select sa$ } & {\footnotesize number of times the action $a$ is selected in state
$s$ (initially, $n(s,a)=0$ for all $s$ and $a$)}\tabularnewline
\hline 
{\footnotesize$\smv s$ } & {\footnotesize value of state $s$ in the game tree}\tabularnewline
\hline 
{\footnotesize$\mv sa$ } & {\footnotesize value obtained after playing action $a$ in state $s$}\tabularnewline
\hline 
{\footnotesize$\svresol s$ } & {\footnotesize completion value of state $s$ ($0$ by default)}\tabularnewline
\hline 
{\footnotesize$\vresol sa$ } & {\footnotesize completion value obtained after playing action $a$
in state $s$}\tabularnewline
\hline 
{\footnotesize$\sresol s$ } & {\footnotesize resolution value of state $s$ ($0$ by default)}\tabularnewline
\hline 
{\footnotesize$\resol sa$ } & {\footnotesize resolution value obtained after playing action $a$
in state $s$}\tabularnewline
\hline 
{\footnotesize$f(s)$ } & {\footnotesize the used evaluation function (point of view of the root
player): $f\left(s\right)=\begin{cases}
\hterminal(s) & \text{if }\terminal s\\
\hadapt(s) & \text{otherwise}
\end{cases}$}\tabularnewline
\hline 
{\footnotesize$\hadapt(s)$ } & {\footnotesize adaptive evaluation function (of non-terminal game tree
leaves ; point of view of the root player)}\tabularnewline
\hline 
{\footnotesize$\hterminal(s)$ } & {\footnotesize evaluation of terminal states, e.g. scoring (point of
view of the root player)}\tabularnewline
\hline 
{\footnotesize$\hfb(s)$ } & {\footnotesize binary evaluation of terminal states, e.g. gain game:
-1 / 0 / 1 (point of view of the root player)}\tabularnewline
\hline 
\end{tabular}}{\footnotesize\par}
\par\end{centering}
\begin{centering}
 
\par\end{centering}
\caption{Index of symbols}
\label{tab:Index-of-symbols} 
\end{table*}

\begin{algorithm}
\begin{centering}
\includegraphics[scale=0.5]{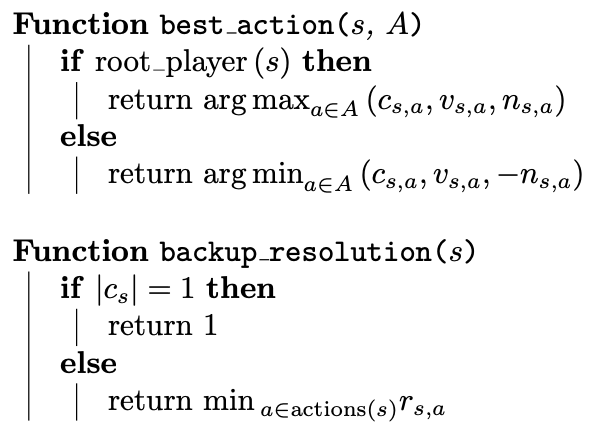}
\par\end{centering}
\caption{Definition of the algorithms best\_action($s$, $A$), which computes
the \emph{a priori }best action, and backup\_resolution($s$), which
updates the resolution of $s$ from its children states (see Table~\ref{tab:Index-of-symbols}
for the definitions of symbols).  }\label{alg:ubfm_best_action_and_backup}

\end{algorithm}

\begin{algorithm}
\begin{centering}
\includegraphics[scale=0.5]{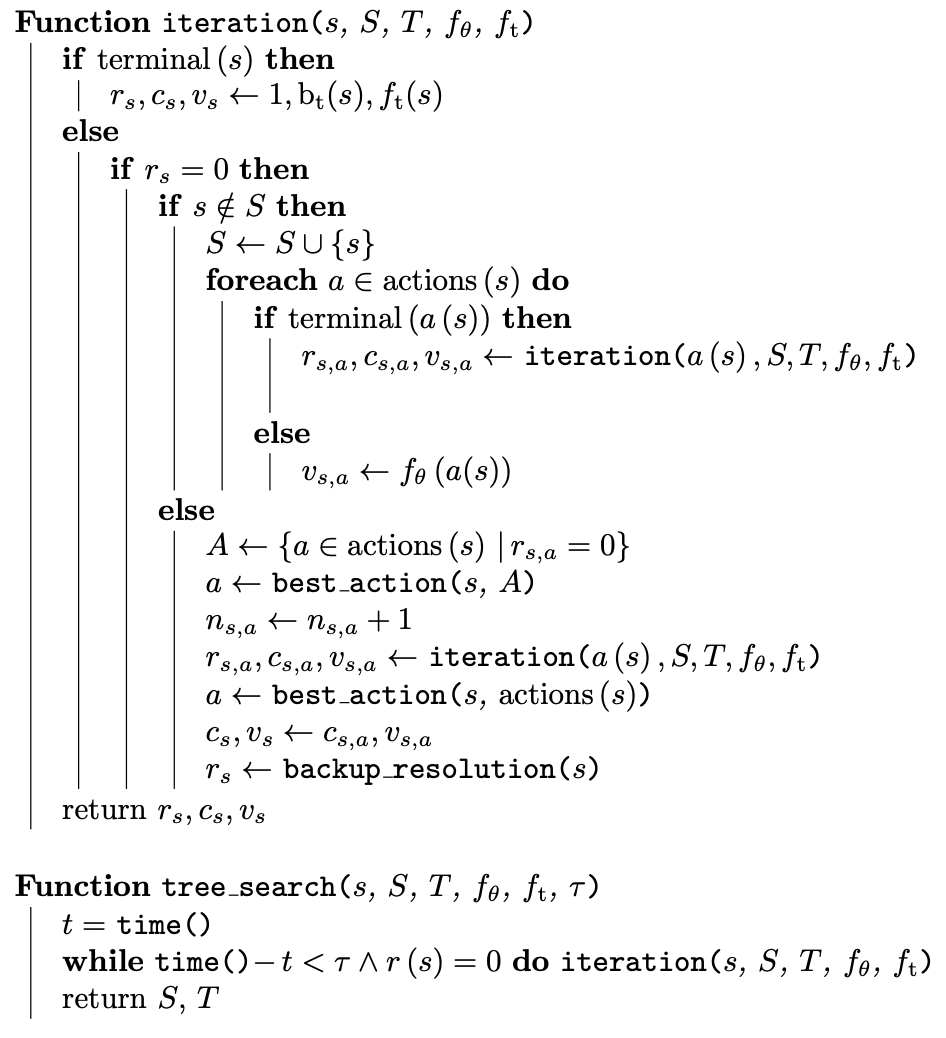}
\par\end{centering}
\caption{Unbounded Minimax tree search algorithm with completion (i.e., including
the calculation of the resolution values c and r of the states ; see
Table~\ref{tab:Index-of-symbols} for symbol definitions and Algorithm~\ref{alg:best_action_and_backup_minibaln}
for best\_action($s$) and backup\_resolution($s$) definitions).
Note: $T=(v,c,r,n)$. }\label{alg:tree_search}

\end{algorithm}

\subsection{Dynamic Difficulty Adjustment}

\label{subsec:Dynamic-Difficulty-Adjustment}

The games research community has a long history of developing mechanisms
to keep human players within the so-called flow window---challenged
but not overwhelmed. Dynamic Difficulty Adjustment (DDA) research
encompasses a wide range of approaches: selecting a static difficulty
level before the game starts \cite{fujita2022alphadda}, adjusting
difficulty parameter multipliers~\cite{sutoyo2015dynamic}, modeling
and clustering players to estimate their skill~\cite{lora2016dynamic,xue2017dynamic},
using domain-dependent parameterized MCTS for balancing~\cite{ishihara2018monte},
and designing online player models that adapt content to maximize
engagement~\cite{xue2017dynamic}. However, these approaches suffer
from at least one of three main limitations~\cite{fujita2022alphadda}: 
\begin{enumerate}
\item reliable player modeling requires information about the opponent (such
as a history of interactions); 
\item DDA systems typically modify game parameters rather than the intrinsic
decision-making process of the agent; and 
\item they rely on domain-dependent techniques. 
\end{enumerate}

\subsection{AlphaDDA}

\label{subsec:AlphaDDA}

The work most closely related to ours is the AlphaDDA algorithm~\cite{fujita2022alphadda}.
AlphaDDA is a general algorithm and a variant of AlphaZero that adapts
its decision-making process to the level of its opponent, notably
during the course of a game. However, it has two main limitations:
first, it is highly parameterized, and this parameterization relies
on grid search, which requires costly tuning; second, it necessitates
a large number of matches against the opponent with whom balanced
play is desired (or against any players of equivalent skill).

AlphaDDA preserves the architecture of AlphaZero --- a deep residual
neural network combined with Monte Carlo Tree Search --- but modifies
the decision-making process so that the agent can dynamically weaken
or strengthen itself according to the estimated value of the current
game state. The core insight is that the value computed by AlphaZero
provides a reasonably accurate estimate of the win probability from
the current board state. By mapping this value to control parameters
of the search or evaluation, the agent can intentionally deviate from
optimal play to better match its opponent’s strength. Three versions
of AlphaDDA have been proposed: 
\begin{itemize}
\item AlphaDDA1 adjusts the number of MCTS simulations based on the estimated
game state value, thereby controlling search depth and effective playing
strength. The number of simulations is given by the following formula: 
\end{itemize}
\[
N_{\mathrm{sim}}\left(v\right)=\min\left(N_{\max},\left\lceil 10^{A_{\mathrm{sim}}\left(\bar{v}+B_{\mathrm{sim0}}\right)}\right\rceil \right)
\]

where $\bar{v}$ is the mean of the estimated values of the last $N_{h}$
game states (from the current player’s perspective), $\left\lceil x\right\rceil $
is the ceiling function, and $A_{\mathrm{sim}}$, $B_{\mathrm{sim0}}$,
$N_{h}$, and $N_{\max}$ are parameters. These values are determined
via grid search to minimize the difference in win and loss rates against
a pool of opponents. 
\begin{itemize}
\item AlphaDDA2 introduces a dropout layer at inference time in the neural
network, with a dropout probability depending on the estimated value.
This reduces the network’s accuracy in a controllable way, weakening
the agent when necessary. 
\item AlphaDDA3 modifies the UCT score used in MCTS so that the agent prefers
suboptimal moves, thereby reducing its likelihood of winning. 
\end{itemize}
In experiments, AlphaDDA1 achieves the best results.

\section{Contribution: Modeling the Balanced Player}

\label{sec:Contribution:-Modeling}

We now turn to the modeling of an artificial player designed to play
in a balanced manner---that is, to challenge its opponents without
either overwhelming them or deliberately conceding victory. Our goal
is to develop a general algorithm that genuinely adapts its strategy
to its opponent---rather than merely adjusting parameters---while
requiring no prior knowledge about the opponent, a property that,
to our knowledge, no existing algorithm in the literature satisfies.

Note that when we refer to a strategy in the context of balancing
concepts, we mean a probability distribution over the available actions.
In contrast, the algorithms we propose operate in the setting of extensive-form
games with deterministic action choices. We denote by $g$ the gain
function (i.e. the payoff, e.g. the score) of the game, taking values
in $\left]-\inf,+\inf\right[$. 

\subsection{First Modeling}

\label{subsec:First-Modeling}

Our first modeling choice is simple and intuitive: the agent’s objective
is to achieve a final outcome as close as possible to a neutral result.
Formally, this corresponds to minimizing the absolute value of the
game outcome at the end of the match: 
\[
\min_{x\in X}\left|\min_{y\in Y}g\left(x,y\right)\right|
\]

where $X$ denotes the set of strategies available to the balanced
player, $Y$ the set of strategies of the opponent, and $g$ the game
outcome (from the balanced player’s perspective) at the end of the
match when the balanced player follows strategy $x$ and the opponent
follows strategy $y$. We refer to this principle of balanced play
as the $\minibaln$ concept (Minimum-Balance-Near).

The challenge then lies in identifying which search algorithm should
be employed to achieve this objective. To this end, we propose an
adaptation of Unbounded Minimax to the balanced-play setting. The
adaptation follows a best-first expansion scheme:

• Balanced player’s turn: the agent selects the action leading to
the successor state with the smallest absolute value.

• Opponent’s turn: the search assumes the opponent acts rationally,
selecting the action that maximizes its own score (equivalently, minimizes
the balanced player’s score).

We call this algorithm Unbounded $\minibaln$. Its overall search
procedure is identical to Unbounded Minimax (see Algorithm~\ref{alg:tree_search}),
but it differs in its two core functions: best\_action() and backup\_resolution()---which
are described in Algorithm~\ref{alg:best_action_and_backup_minibaln}.
Method best\_action() applies the optimization of the chosen action
following the minibal concept and backup\_resolution() computes the
resolution of states: a state where the balanced player acts is resolved
if it has a resolved draw action or if all its children are resolved. 

\begin{algorithm}
\begin{centering}
\includegraphics[scale=0.5]{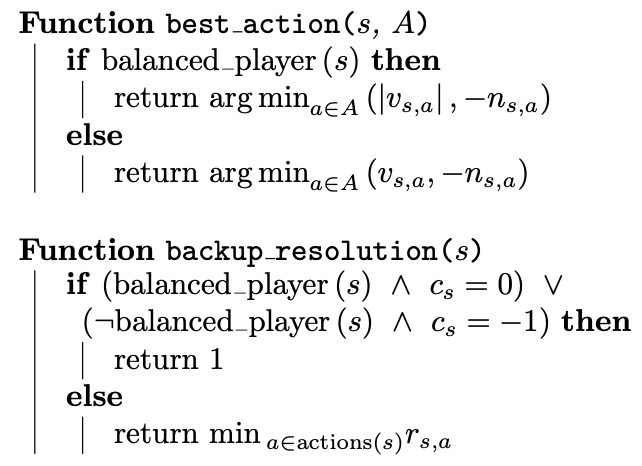}
\par\end{centering}
\caption{Definition of the following algorithms for Unbounded $\protect\minibaln$:
best\_action($s$, $A$), which computes the \textsl{a priori} best
action using completion, and backup\_resolution($s$), which updates
the resolution of $s$ based on its child states. }\label{alg:best_action_and_backup_minibaln}

\end{algorithm}

\subsection{Second Modeling}

\label{subsec:Second-Modeling}

As we will show in the experimental section, the first algorithm performs
poorly---it loses far too often. We therefore adopt a different modeling
approach.

This time, the objective is no longer to obtain a final score as close
as possible to zero. Instead, the goal is to achieve a positive score
that is as close to zero as possible. If such a result cannot be attained,
the agent aims for a negative score that is as close to zero as possible.
Formally: 
\[
\min_{x\in X}\left(\indicatrice{\mathbb{R}_{*}^{-}}{\min_{y\in Y}g\left(x,y\right)},\left|\min_{y\in Y}g\left(x,y\right)\right|\right)
\]

where tuple minimization is performed lexicographically, $\indicatrice{\mathbb{R}_{*}^{-}}z=\begin{cases}
1 & \text{if }z<0\\
0 & \text{otherwise}
\end{cases}$, $X$ denotes the set of strategies of the balanced player, $Y$
the set of strategies of the opponent, and $g$ the game outcome (from
the balanced player’s perspective) at the end of the match when the
balanced player follows strategy $x$ and the opponent follows strategy
$y$.

We refer to this principle as the $\minibalp$ concept (Minimum-Balance-Positive).

To implement this concept, we propose a corresponding adaptation of
Unbounded Minimax, which we call Unbounded $\minibalp$. Its overall
search algorithm is identical to that of Unbounded $\minibaln$ (i.e.,
Algorithm~\ref{alg:tree_search} with backup\_resolution() defined
in Algorithm~\ref{alg:best_action_and_backup_minibaln}), but the
best\_action() function is modified and defined in Algorithm~\ref{alg:best_action_and_backup_minibalp}.

\begin{algorithm}
\begin{centering}
\includegraphics[scale=0.5]{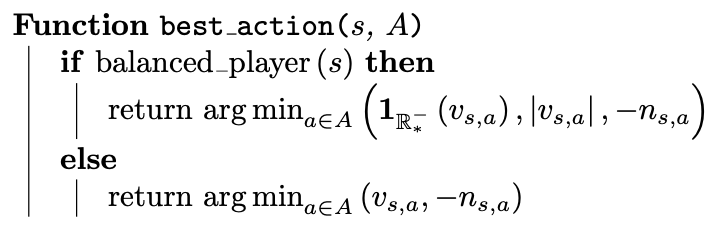}
\par\end{centering}
\caption{Definition of best\_action($s$, $A$) for Unbounded $\protect\minibalp$,
which computes the \emph{a priori }best action. }
\label{alg:best_action_and_backup_minibalp} 

\end{algorithm}

The only difference from Unbounded $\minibaln$ lies in the balanced
player’s decision rule:

• If at least one successor state has a positive evaluation, the agent
selects the action leading to the smallest positive value.

• If no such state exists, the agent selects the action leading to
the largest (i.e., least negative) evaluation.

\section{Experiments}

\label{sec:Experiments}

We now present the experiments conducted to evaluate the contributions
of this paper. Our primary objective is to assess the ability of our
three algorithms to balance matches. To this end, we design a series
of matches involving both high-level neural evaluation functions (trained
via reinforcement learning) and low-level neural evaluation functions
(also trained via reinforcement learning). More specifically, our
aim is to compare how effectively algorithms using strong neural networks
can balance against weaker networks.

We analyze the results from two complementary perspectives. First,
we examine the win rate, where a rate close to 50\% indicates balanced
play. Second, we evaluate the average minimax value of the final states
of the matches, with an average value of 0 corresponding to balanced
outcomes.

\subsection{Evaluation Protocol}

\label{subsec:Evaluation-Protocol}

We evaluate Unbounded $\minibaln$, Unbounded $\minibalp$, and classical
Unbounded Minimax as a reference.

The first performance metric, \textsl{binary gain}, corresponds to
the average game outcome: a victory counts as $+1$ and a defeat as
$-1$ (this metric is more informative than the win rate for games
that allow draws). The gain is averaged across all matches.

The second performance metric, \textsl{score}, corresponds to the
average terminal evaluation: $\frac{1}{\left|S\right|}\sum_{s\in S}\fterminal\left(s\right)$,
where $S$ denotes the set of endgame states reached in the performed
matches, and $\fterminal$ is the terminal evaluation function used
by the search algorithms. 

For each performance metric, the best balancing algorithm is the one
whose average evaluation is closest to zero, indicating superior adaptation
to its opponent.

Each evaluated algorithm employs a high-level evaluation function
and plays against Unbounded Minimax equipped with a low-level evaluation
function. Performance is averaged over all high-level/low-level evaluation
function pairs.

Experiments are conducted across seven games: International Draughts,
Chess, Lines of Action, Connect6, Outer-Open-Gomoku, Xiangqi, and
Havannah. Overall, results are averaged across all these games. 

\subsection{Technical Details}

For the balanced player, we evaluate multiple search budgets to study
how binary gain and score evolve as a function of this parameter.

For each game, we trained 40 state evaluation functions via reinforcement
learning using the Descent Minimax algorithm from the Athénan framework,
over a total of 70 days of training.

• The 20 strongest evaluation functions constitute the set of high-level
evaluations.

• The 20 weakest evaluation functions constitute the set of low-level
evaluations.

To further amplify the difference in level, the low-level evaluation
functions were reverted to their 20-day checkpoint rather than using
their final 70-day versions. 

For each evaluated search time and performance metric, an algorithm’s
performance is measured over 800 matches per game (20 high-level evaluation
functions against 20 low-level evaluation functions, considering both
first and second player positions). 

More technical details are provided in the Appendix.

\subsection{Results}

\label{subsec:Results}

We now present the performance of each algorithm according to the
two performance metrics: binary gain and score.

\begin{figure}
\begin{centering}
\includegraphics[scale=0.33]{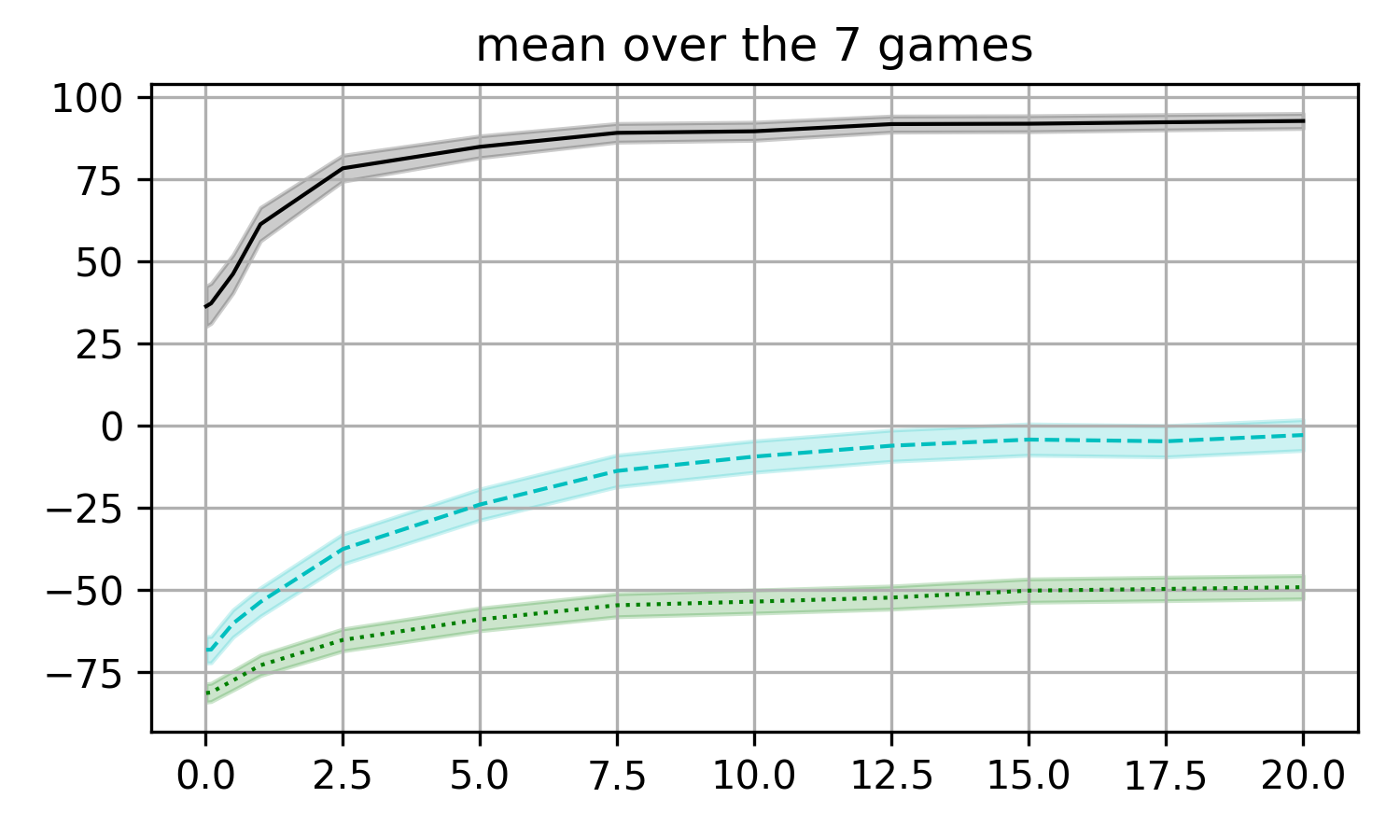} \includegraphics[scale=0.33]{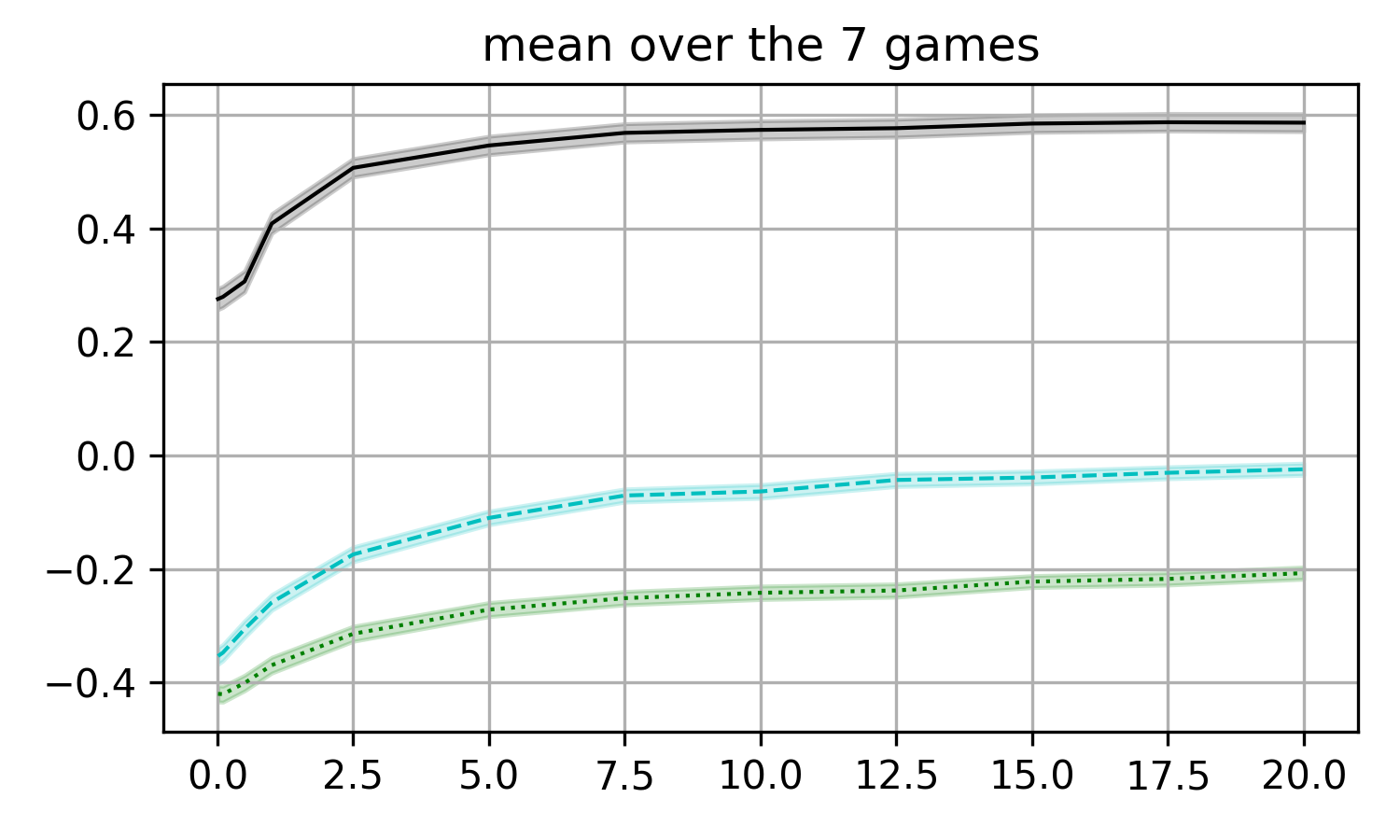} 
\par\end{centering}
\centering{}\caption{Average binary gain (left) and average scores (right) as a function
of search time (in seconds) for \textcolor{teal}{$\protect\minibaln$
(dotted green curve)}, \textcolor{cyan}{$\protect\minibalp$ (dashed
cyan curve)}, and \textcolor{gray}{Unbounded Minimax (black line)}
across the studied games (in percentage).}
\label{fig:courbe_moyenne} \label{fig:courbe_moyenne-score} 
\end{figure}

The average binary gains (resp. average scores) are shown in Figure~\ref{fig:courbe_moyenne}
(resp. in Figure~\ref{fig:courbe_moyenne-score}), which presents
the outcomes as a function of search time per move. The average binary
gains (resp. average scores) for a search time of 20 seconds per move
are reported in Table~\ref{tab:The-average-binary} (resp. in Table~\ref{tab:The-average-scores}).
Detailed results for each individual game are provided in Appendix.

\begin{table}
\centering{}{\tiny{}%
\begin{tabular}{|c|c|c|c|c|c|c|c|}
\hline 
{\tiny Unbounded } & {\tiny gain } & {\tiny 95\% C.R } & {\tiny win } & {\tiny draw } & {\tiny loss} & {\tiny score} & {\tiny 95\% C.R }\tabularnewline
\hline 
\hline 
{\tiny Minimax } & {\tiny 92.68 } & {\tiny 0.91} & {\tiny 94.77} & {\tiny 3.14} & {\tiny 2.09} & {\tiny 0.59} & {\tiny 0.016}\tabularnewline
\hline 
{\tiny$\minibalp$ } & {\tiny -2.96} & {\tiny 1.95} & {\tiny 21.45} & {\tiny 54.14} & {\tiny 24.41} & {\tiny -0.024} & {\tiny 0.011}\tabularnewline
\hline 
{\tiny$\minibaln$ } & {\tiny -68.95} & {\tiny 1.24} & {\tiny 1.19} & {\tiny 30.05} & {\tiny 69.45} & {\tiny -0.21} & {\tiny 0.011}\tabularnewline
\hline 
\end{tabular}}{\tiny}\caption{Average binary gains and win, draw, loss rates in percentage and scores
at 20 seconds of search time per move across the seven games.}
\label{tab:The-average-binary} \label{tab:The-average-scores} 
\end{table}

As expected, the classical Unbounded Minimax achieves a high win rate
and scores, confirming that the high-level evaluation functions are
indeed much stronger than the low-level ones. However, Unbounded $\minibaln$
yields surprisingly low gains and scores (except in two games where
its performance is similar to that of other balance algorithms). This
suggests that $\minibaln$ is not well suited for balanced play.

Most notably, Unbounded $\minibalp$ consistently achieves the best
results across all games. Its gains and scores are the closest to
zero among all tested algorithms, and in most cases, it reaches nearly
perfect balance.

\subsubsection{Interpretation}

These findings demonstrate that, among the new algorithms introduced
in this paper, $\minibalp$ appears to be a practical and effective
approach for achieving balanced play, sometimes even managing to produce
perfectly balanced matches.  $\minibaln$, on the other hand, behaves
like a player who loses very frequently, yet almost always by a narrow
margin.

\section{Discussion}\label{sec:Discussion}

\subsection{Very Weak Opponents}

We also repeated the experiments of this paper against extremely weak
opponents by replacing the weak baselines with a standard MCTS player
(i.e., MCTS with UCT and no evaluation function; see Appendix for
details). The results show that when the skill gap between players
is very large, the balancing algorithm cannot achieve perfect balancing
within a reasonable computation time. This limitation is not unexpected:
similarly to algorithms designed to maximize winning probability,
it is generally impossible to guarantee an optimal balancing strategy
under tight time constraints due to the size of the search space.
For example, in this additional experiment,$\minibalp$ achieves an
average binary gain of $37.95%\ensuremath{}
$ against MCTS with a 20-second search time. Nevertheless, the balancing
algorithm consistently and substantially reduce the initial skill
gap. Indeed, Unbounded Minimax achieves an average binary gain of
$100%\%
$ against MCTS. These results suggest that balancing algorithms should
be viewed as practical, ready-to-use operators whose purpose is to
significantly narrow the strength gap between players, rather than
to guarantee perfect balance in extreme cases. Importantly, we recall
that they operate without any domain-specific knowledge, parameter
tuning, or learning phase.

Thus, the proposed balancing algorithm remains effective even against
very weak opponents. Although perfect balance cannot always be achieved,
increasing the degree of balance is sufficient to improve the user
experience and therefore justifies its practical use. Moreover, $\minibalp$
can be combined with other balancing mechanisms, in particular matchmaking
approaches, to further enhance balance. For instance, a matchmaking
algorithm may first select an evaluation function whose playing strength
is closest to that of the user. Then, $\minibalp$ can be applied
on top of this selection to further narrow the remaining gap and adapt
the opponent more precisely to the user’s skill level.

Finally, it is worth noting that a very simple yet parametric method
can also be used to obtain perfect balanced gameplay against very
weak opponents with $\minibalp$, namely by adjusting the search time
(see the experiment reported in Appendix). As the search time increases,
the average binary gain monotonically increases and eventually reaches
zero, at which point perfect balancing is achieved. If the opponent
is substantially weaker, further increasing the search time leads
to a positive average gain, thereby degrading the balancing performance
as the outcome moves away from zero, until it stabilizes around a
fixed value. Consequently, selecting an appropriate search time is
sufficient to achieve a zero average binary gain in this setting.
Importantly, this method does not hold for Unbounded Minimax. In the
experiment against MCTS, Unbounded Minimax already achieves a win
rate of 99.64\% with only 0.01 seconds of the search time, leaving
no practical parameter range to recover balance. This contrast further
highlights the usefulness of our approach, even when facing extremely
weak opponents.

\subsection{Reflections on User Experience}

We have shown that these operators not only reduce the win-rate gap
against an unknown weaker opponent without relying on any learning
procedure, but also lead to significantly closer endgames. As a result,
the matches produced by these algorithms are inherently more engaging
for human players. More specifically, compared to win-oriented algorithms,
our approach deliberately reduces playing strength, alleviates pressure
on the opponent, and, as directly evidenced by our experiments, increases
the user’s probability of winning.

By losing by smaller margins and winning more frequently, a human
player is more likely to remain engaged and to play additional games,
while being exposed to situations that are more favorable for learning.
Indeed, facing an opponent perceived as impossible to defeat discourages
exploration of alternative strategies, hinders progress, and ultimately
suppresses the desire to replay. Importantly, the win rates achieved
by our balancing algorithms remain sufficiently high to ensure challenging
and meaningful interactions.

A remaining question is whether these algorithms may deliberately
produce artificial defeats. For $\minibaln$, this behavior is both
experimentally confirmed and directly follows from its definition:
since it explicitly aims to minimize the absolute final game outcome,
it may prefer a near-draw defeat over a crushing victory. 

By contrast, $\minibalp$ cannot intentionally concede wins by design.
It first aims to win and, only when it believes it will win, it attempts
to minimize the margin of victory. However, when faced with a choice
between a strongly winning solved action and an action that it evaluates
as only weakly winning, $\minibalp$ will select the latter. This
behavior can sometimes lead to an eventual loss that could theoretically
be perceived as artificial, but only when the strongly winning action
is trivially winning.

This potential issue can be addressed by a simple micro-adjustment
by applying the following rule: always selecting a solved winning
action whenever one is available. While this modification naturally
increases the win rate, additional experiments (not reported in this
paper) show that its impact on the average binary gain remains slightly
moderate, increasing it from \textminus 2.96\% to 9.82\% in the experimental
setting considered here. This indicates that a strong level of balance
is preserved, even though it is no longer nearly perfect.

Moreover, this adjustment involves playing non-trivial resolved winning
states, which, if omitted, would not have been perceived as artificial
wins by human players. A finer improvement can therefore be achieved
by restricting this rule to wins resolved to a depth of at most $d$.
Finally, as previously noted, any excess gain introduced by such modifications
can, if necessary, be compensated by a reduction in thinking time.

Finally, note that a rigorous statistical study in which balancing
algorithms face human players, in order to verify that no other problematic
cases arise, is a research perspective that is beyond our current
means.

\section{Conclusion}

\label{sec:Conclusion}

\subsubsection*{Summary}

In this paper, we addressed the problem of balanced play, that is,
designing agents capable of challenging their opponents without either
overwhelming them or conceding artificially. We first proposed two
conceptual models of balance, then introduced the corresponding algorithms
--- Unbounded $\minibaln$ and Unbounded $\minibalp$ --- specifically
designed to enable artificial agents to play in a balanced manner.

These algorithms contrast, on the one hand, with classical methods
whose sole objective is winning, such as Unbounded Minimax. On the
other hand, they stand in opposition to Dynamic Difficulty Adjustment
approaches in the literature, which either rely on domain-dependent
heuristics, require prior knowledge of the opponent to achieve balance,
or do not adapt their strategy dynamically to the opponent.

Our experimental evaluation across seven board games demonstrated
that Unbounded $\minibalp$is capable of producing outcomes closest
to zero gain. Remarkably, in many settings, it achieved nearly perfect
balance, providing compelling evidence that it represents a significant
step forward in the design of game-playing agents adapted to human
needs. By contrast, Unbounded $\minibaln$ exhibited unexpectedly
poor performance, with a disproportionately high loss rate, making
it unsuitable for balanced play. Nevertheless, it appears to be a
promising candidate for scenarios where an opponent that loses frequently
, but only by narrow margins, is desirable. 

Overall, these results show that $\minibalp$ currently constitutes
the most effective approach for producing balanced agents without
online learning or prior knowledge of the opponent. This contribution
opens the way to new applications not only in entertainment and video
games but also in serious games and educational contexts, where maintaining
user engagement without discouragement is essential.

\subsubsection*{Future work}

Several avenues for future work naturally follow from this study.
First, it is crucial to investigate the causes of the high loss rate
observed for $\minibaln$. Our working hypothesis is that this strategy
too frequently selects marginally losing states; in practice, this
would provide many opportunities for weaker opponents to convert matches
into wins, driving the balanced player’s win rate well below 50\%.
A targeted analysis is therefore required to validate this hypothesis
and identify remedies.

Beyond that, further research should explore extensions to imperfect-information
and stochastic games. The design of terminal evaluation functions
adapted to the balancing objective also requires in-depth study. Finally,
large-scale evaluations involving human players are needed to measure,
in addition to statistical balance, the effects on player engagement,
learning outcomes, and long-term retention in both entertainment and
serious-game contexts.

\bibliographystyle{plain}
\bibliography{jeux}

\section{Appendix}

Section~\ref{subsec:Technical-Details} provides details about the
experiment described in the article. Section~\ref{subsec:Details-Results}
details the results of the experiment described in the article. Section~\ref{subsec:Additional-Experience-Against}
adds an additional experiment against a very weak opponent. Finally,
Section~\ref{subsec:More-Discussion} adds further discussion of
the algorithms.

\subsection{More Technical Details}\label{subsec:Technical-Details}

We will now present the missing technical details of the experiments.

The search time for the low-level player is fixed at 1 second per
move. 

To determine the strongest and weakest evaluation, the evaluation
functions were ranked by strength according to the results of a round-robin
tournament (using Unbounded Minimax as the search algorithm with 1
second per move).

We verified that using 20-day checkpoint rather than using their final
70-day versions downgrade effectively reduced their strength. 

The number of matches for evaluate each algorithm across all seven
games is 5,600 for each evaluated search time. 

The search times considered were: $\left\{ 0.01,0.1,0.5,1,2.5,5,7.5,10,12.5,15,17.5,20\right\} $
seconds per move.

All experiments were conducted on the Adastra supercomputer. Each
compute node consists of one AMD Trento EPYC 7A53 processor (64 cores,
2.0 GHz), 256 GiB DDR4-3200 MHz RAM, 4 Slingshot 200 Gb/s NICs, and
8 GPU devices (4 AMD MI250X accelerators, each comprising 2 GPUs),
with a total of 512 GiB HBM2 memory. All programs were implemented
in Python using TensorFlow.

The specific terminal functions used for each game are listed in Table~\ref{tab:terminal_evaluations}.

\begin{table}
\begin{centering}
\begin{tabular}{|c|c|}
\hline 
 & evaluation\tabularnewline
\hline 
\hline 
Chess & summed piece values\tabularnewline
\hline 
Xiangqi & summed piece values\tabularnewline
\hline 
International Draughts & summed piece values\tabularnewline
\hline 
Lines of Action & additive depth heuristic\tabularnewline
\hline 
Havannah & multiplicative depth heuristic\tabularnewline
\hline 
Outer-Open-Gomoku & multiplicative depth heuristic\tabularnewline
\hline 
Connect6 & multiplicative depth heuristic\tabularnewline
\hline 
\end{tabular}
\par\end{centering}
\caption{Terminal evaluation functions used for each game (see the Athénan
paper for their definitions).}
\label{tab:terminal_evaluations}
\end{table}

The terminal evaluation functions have been normalized to be in $[-1,1]$

\subsection{Details Results}\label{subsec:Details-Results}

\begin{figure}
\begin{centering}
\includegraphics[scale=0.4]{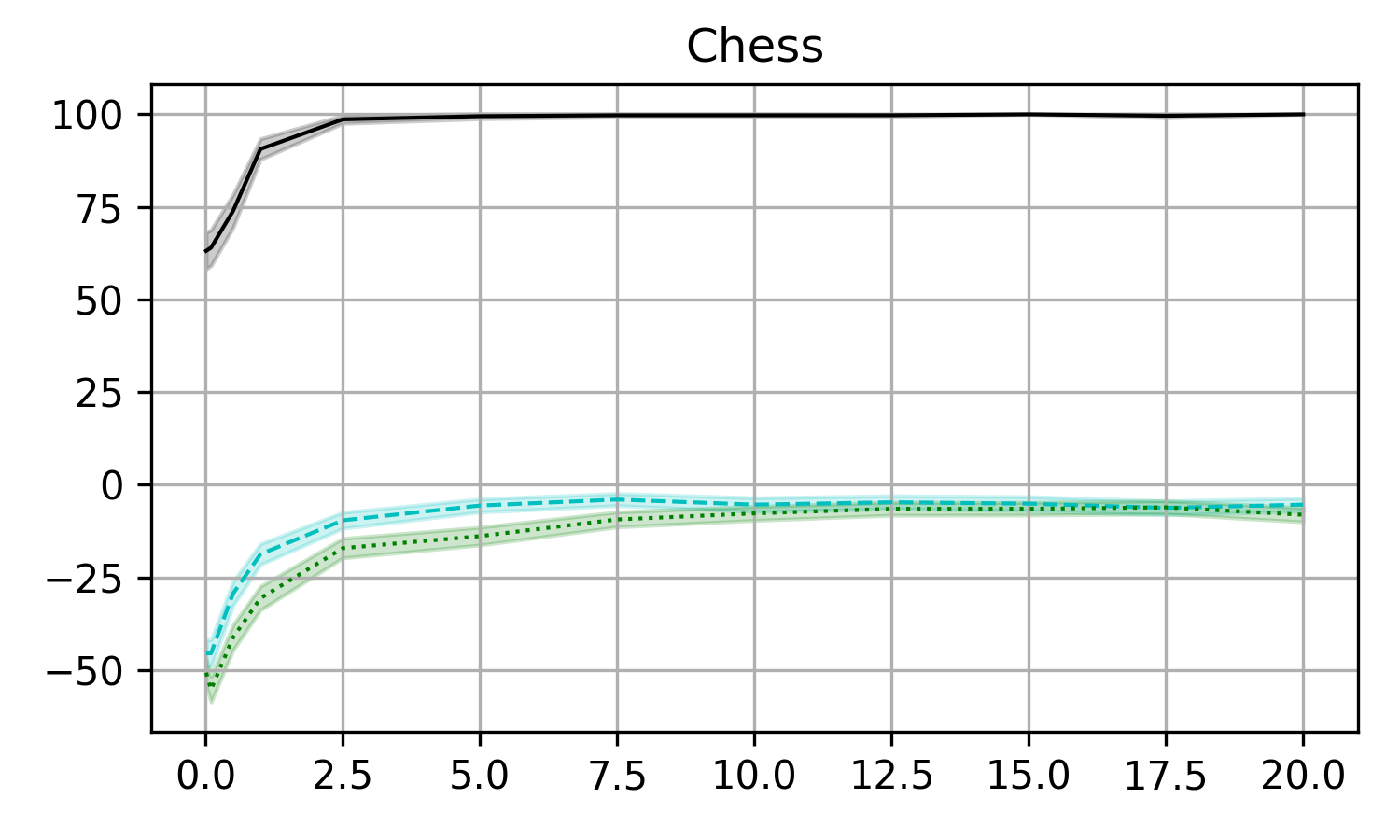}
\par\end{centering}
\begin{centering}
\includegraphics[scale=0.4]{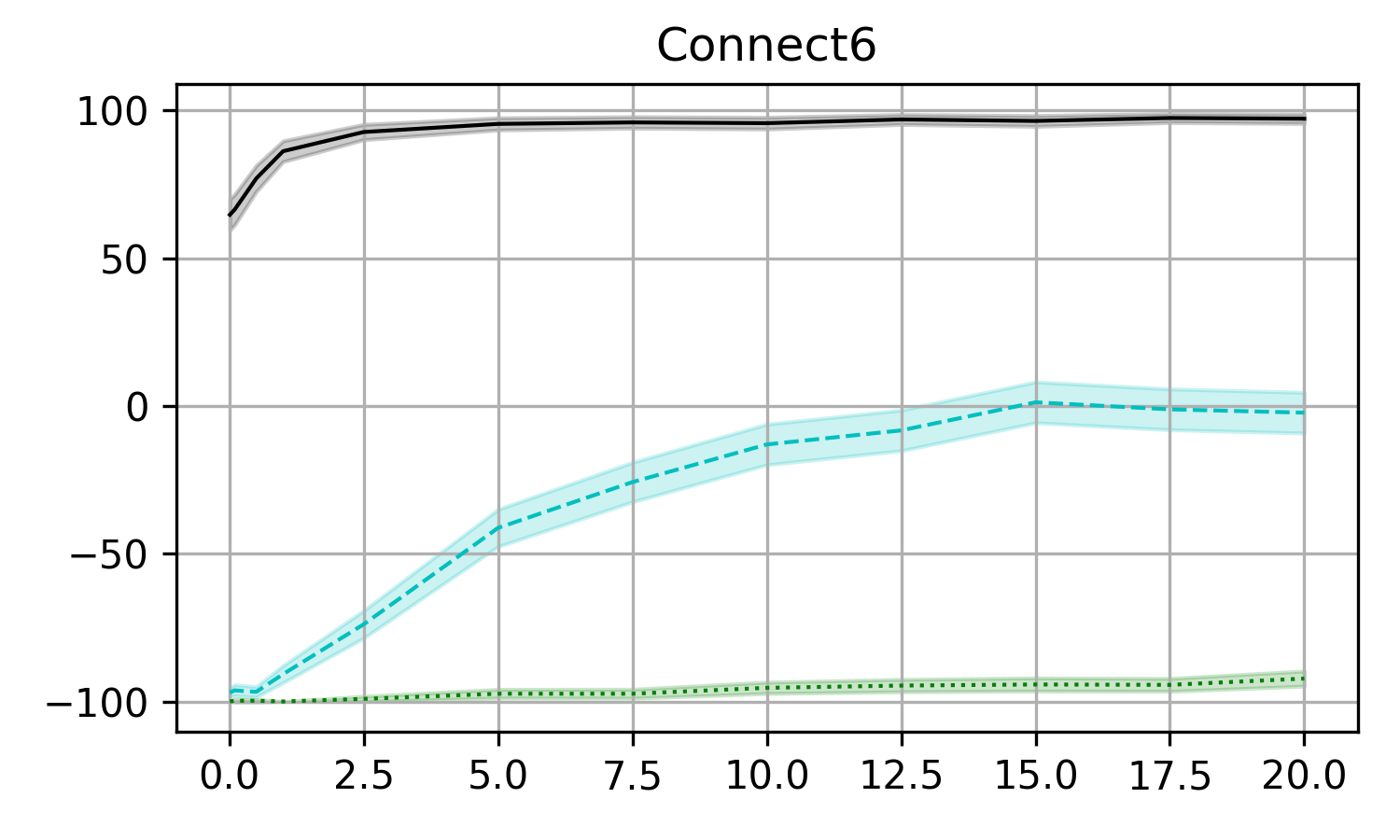}
\par\end{centering}
\begin{centering}
\includegraphics[scale=0.4]{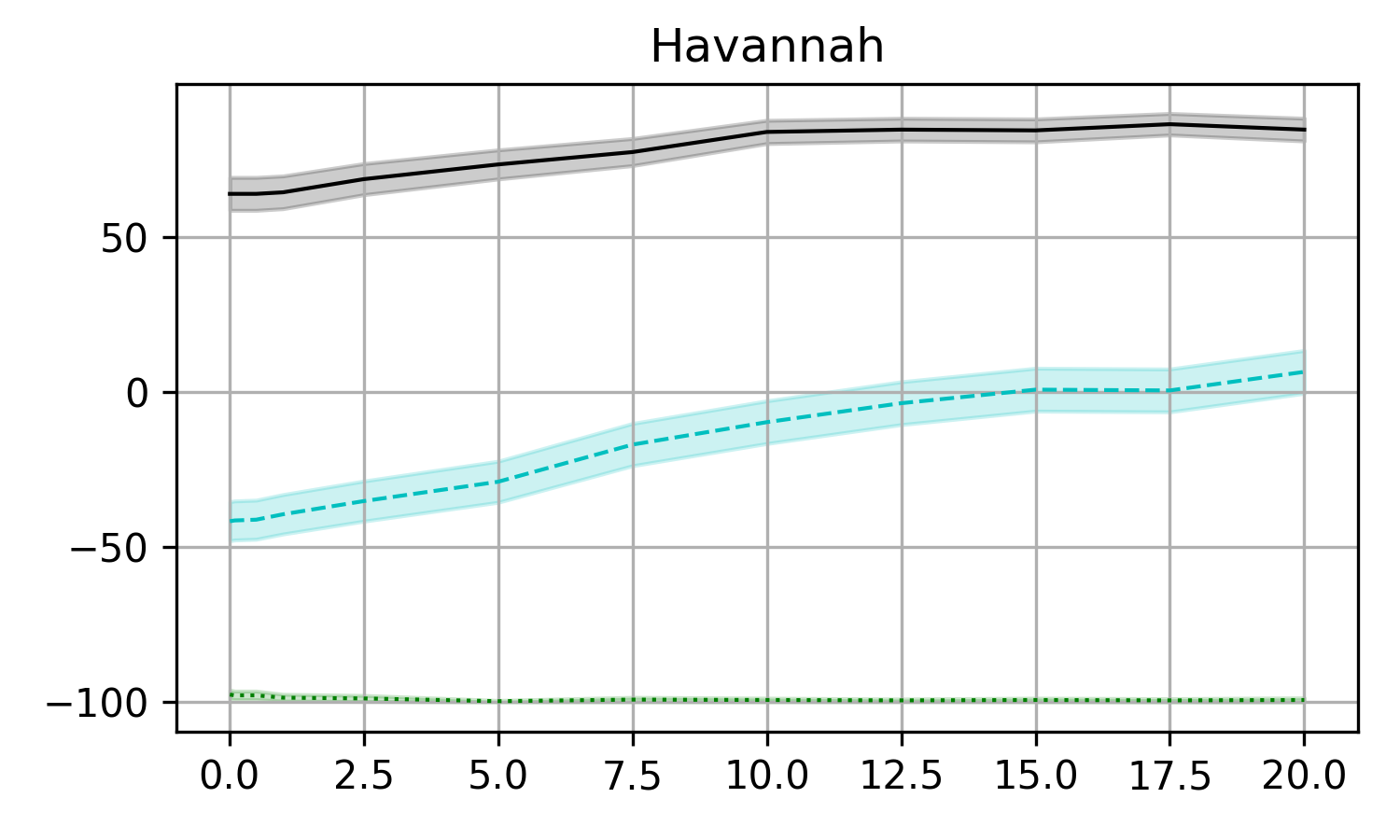}
\par\end{centering}
\begin{centering}
\includegraphics[scale=0.4]{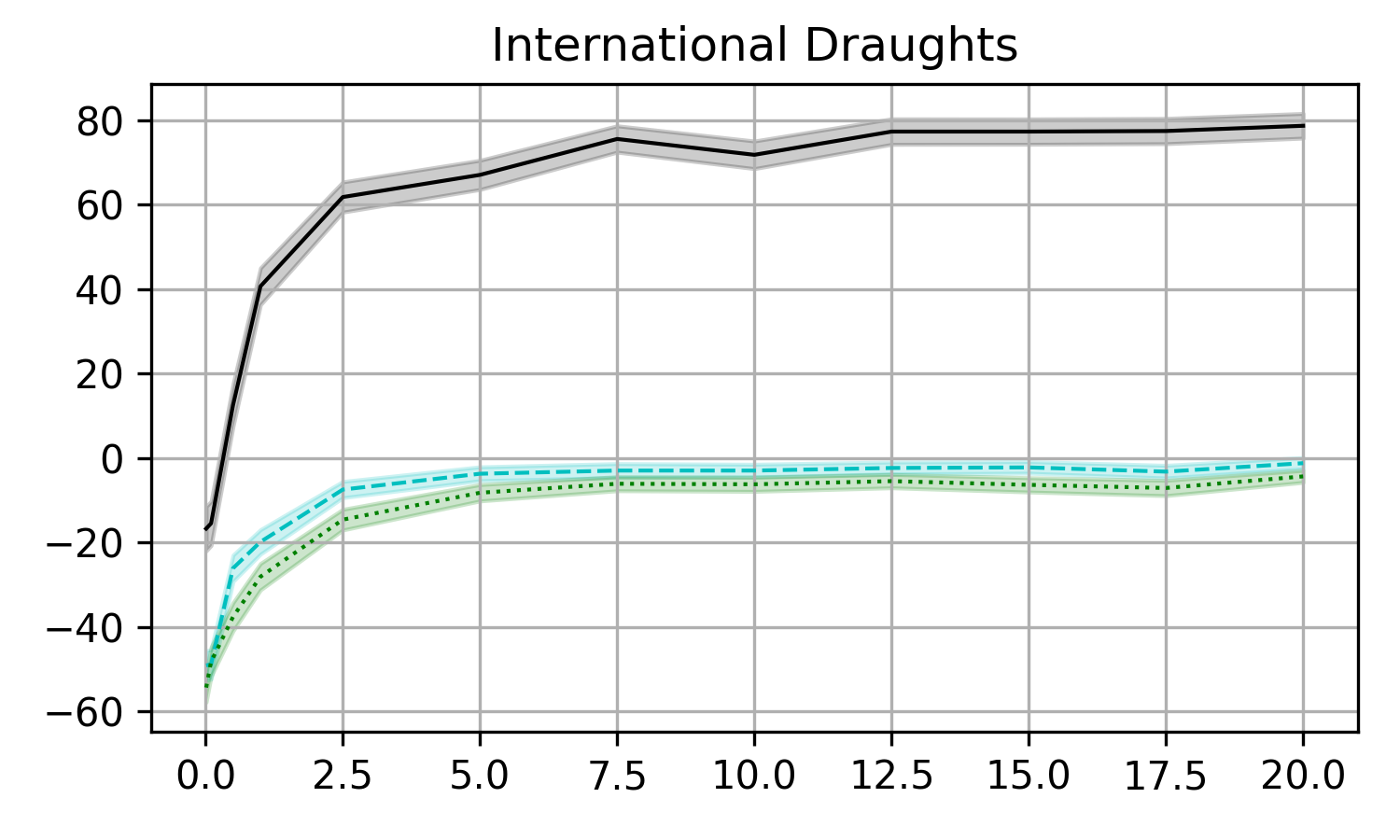} 
\par\end{centering}
\begin{centering}
\includegraphics[scale=0.4]{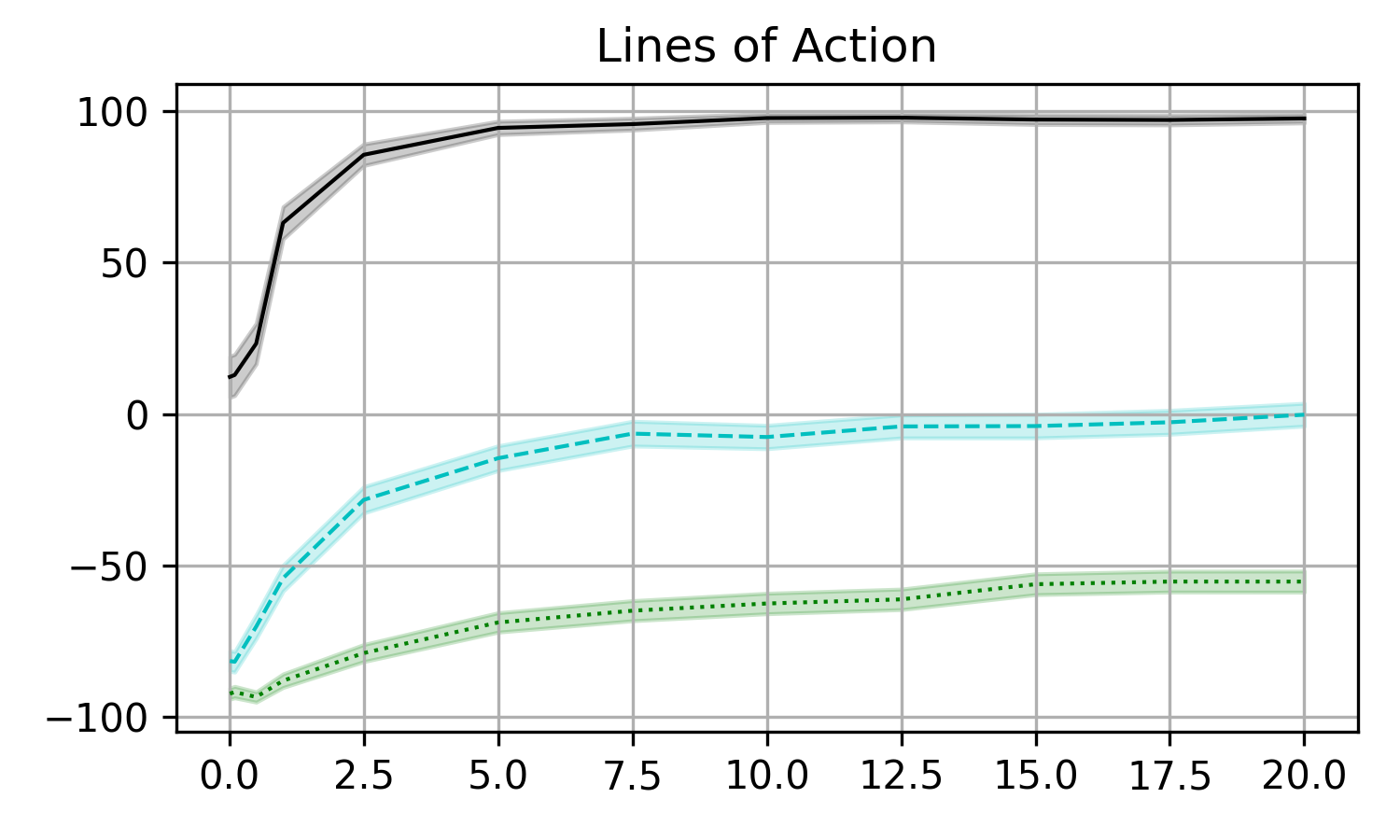}
\par\end{centering}
\begin{centering}
\includegraphics[scale=0.4]{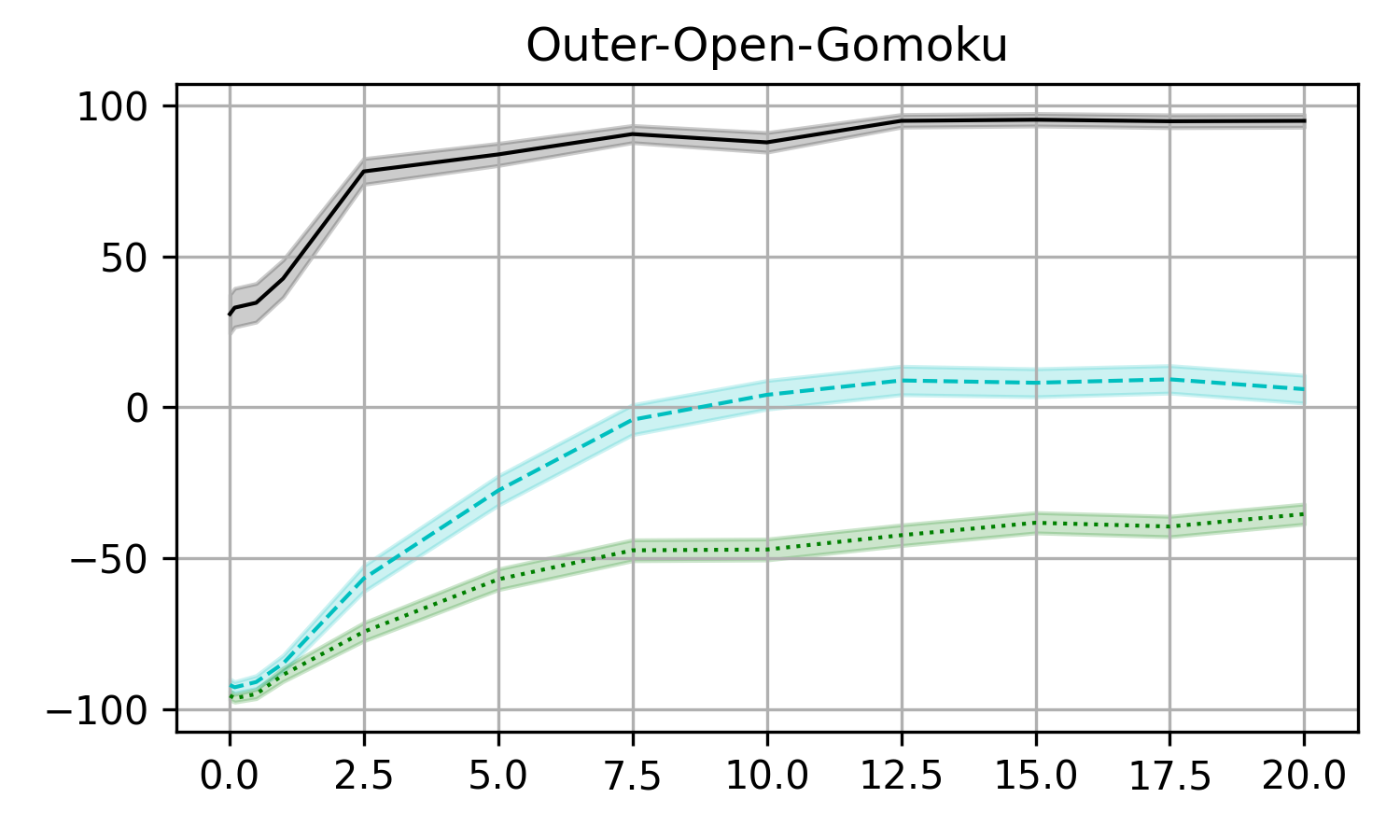}
\par\end{centering}
\begin{centering}
\includegraphics[scale=0.4]{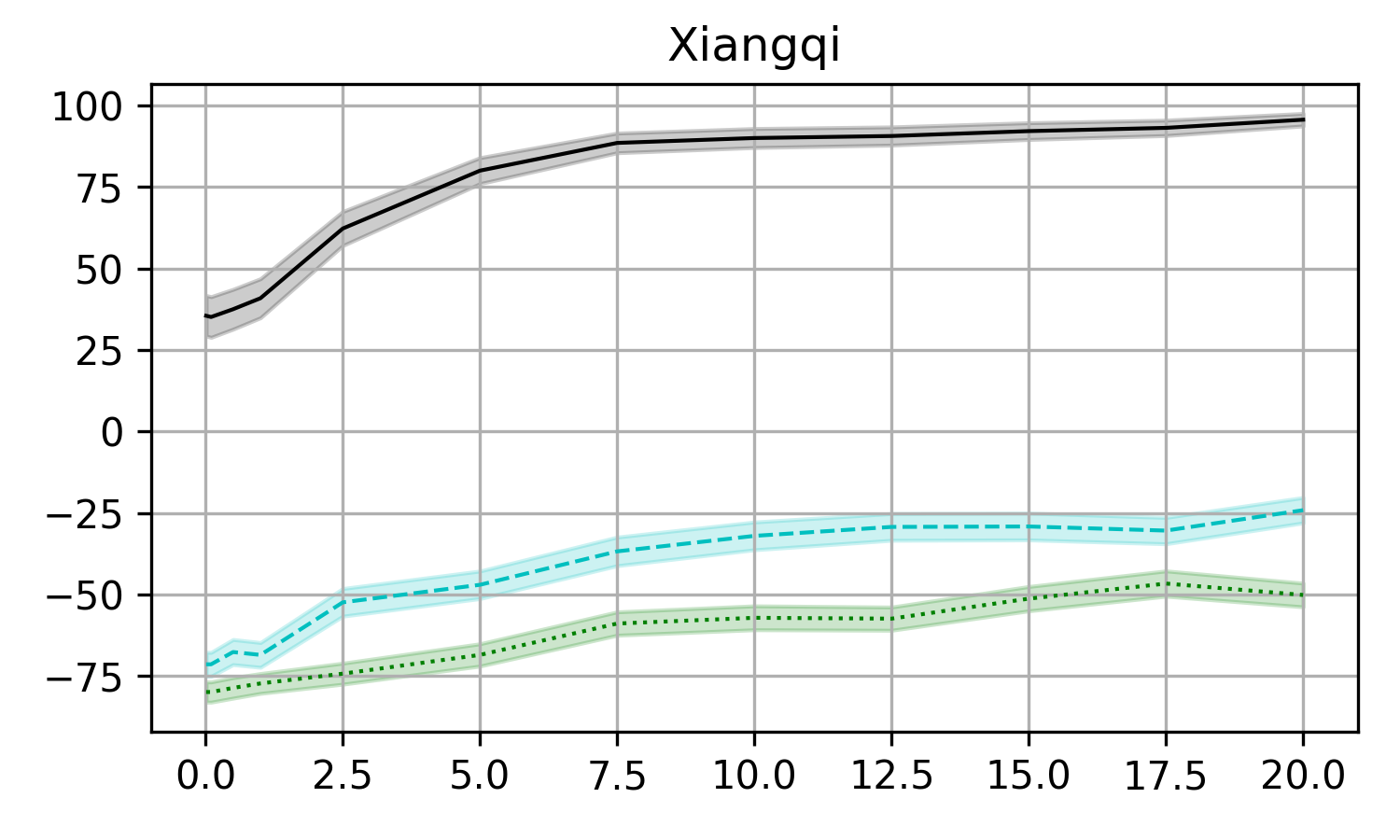}
\par\end{centering}
\caption{Average binary gains as a function of search time (in seconds) for
\textcolor{teal}{$\protect\minibaln$ (dotted green curve)}, \textcolor{cyan}{$\protect\minibalp$
(dashed cyan curve)}, and \textcolor{gray}{Unbounded Minimax (black
line)} across the studied games (C.R.: confidence radius).}
\label{fig:courbe_moyenne-details}
\end{figure}

\begin{figure}
\begin{centering}
\includegraphics[scale=0.4]{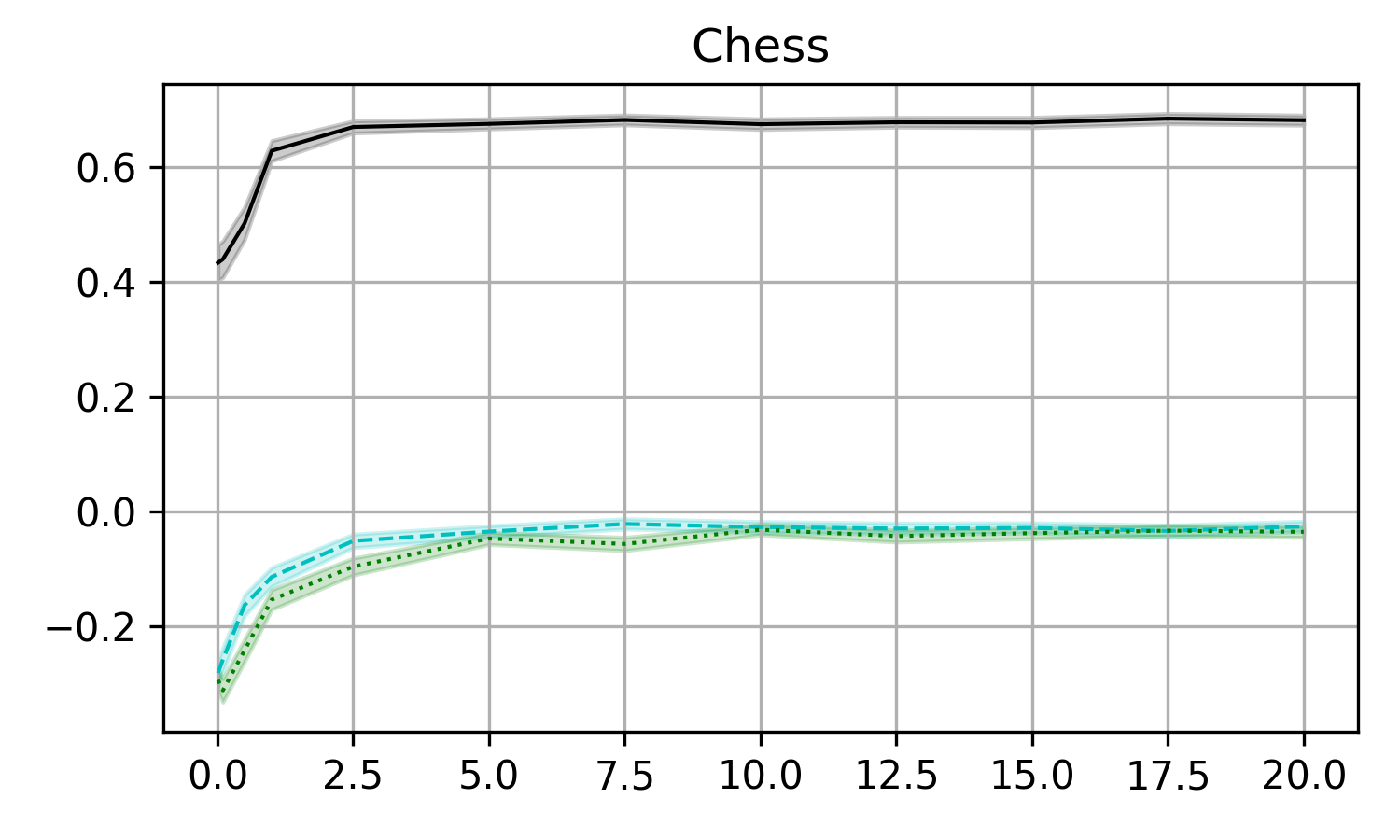}
\par\end{centering}
\begin{centering}
\includegraphics[scale=0.4]{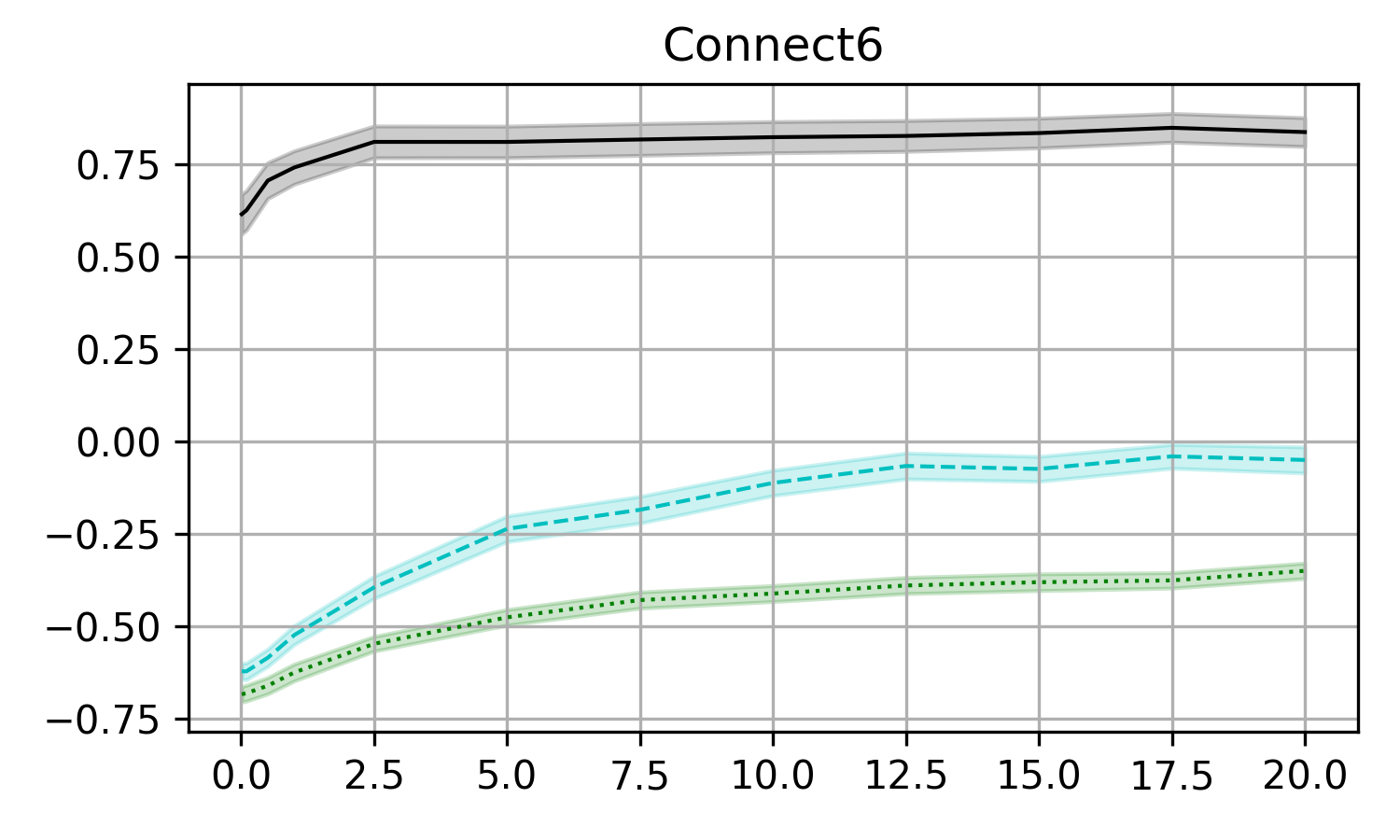}
\par\end{centering}
\begin{centering}
\includegraphics[scale=0.4]{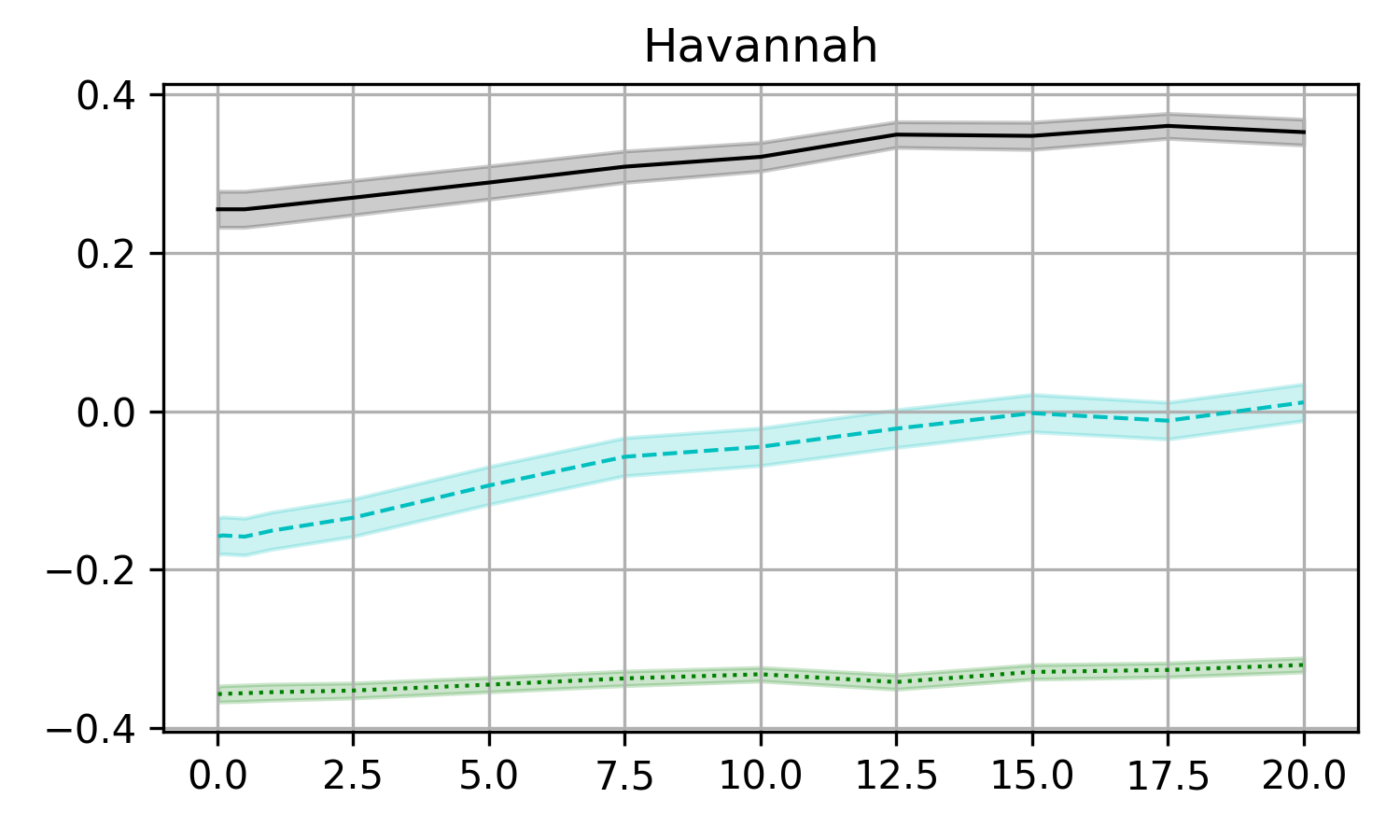}
\par\end{centering}
\begin{centering}
\includegraphics[scale=0.4]{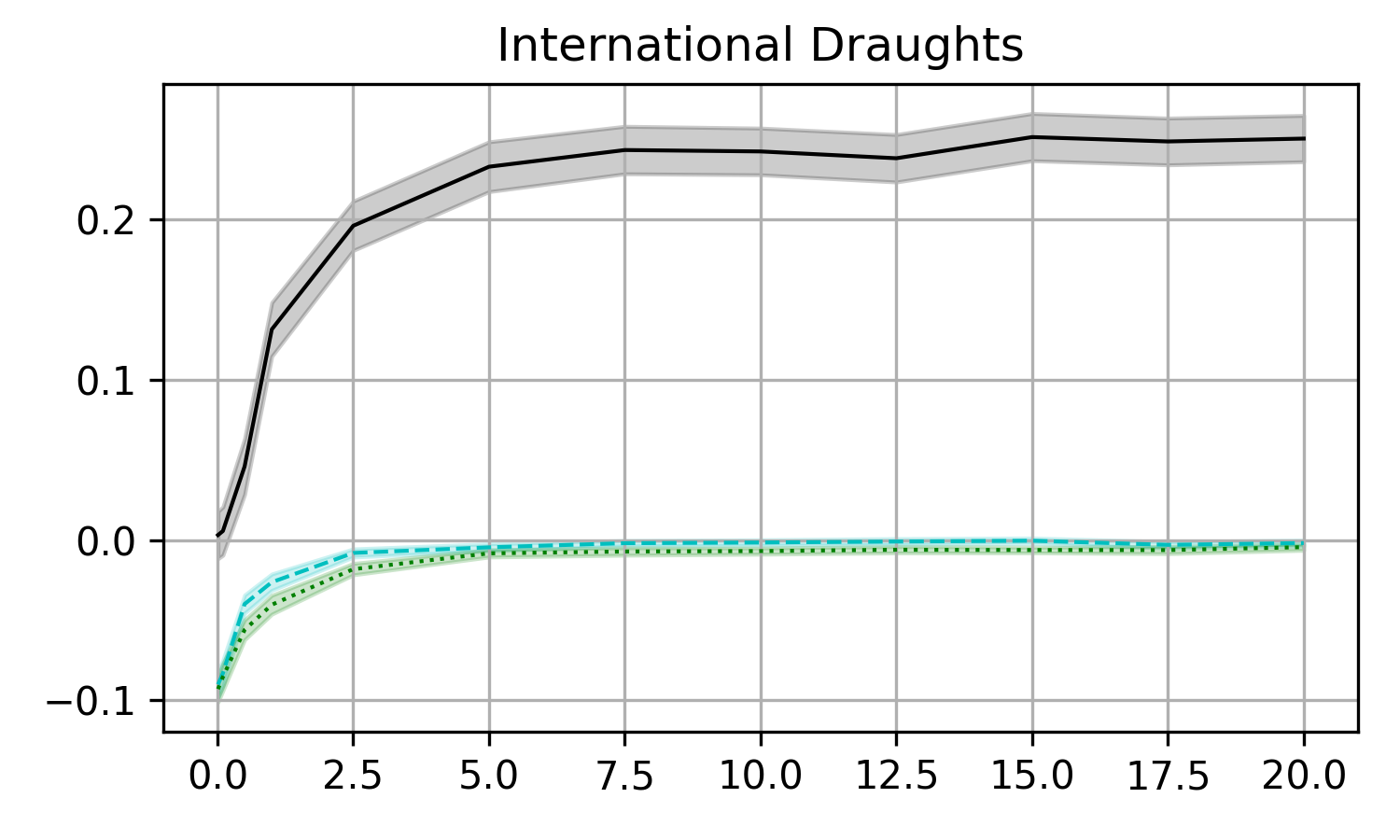} 
\par\end{centering}
\begin{centering}
\includegraphics[scale=0.4]{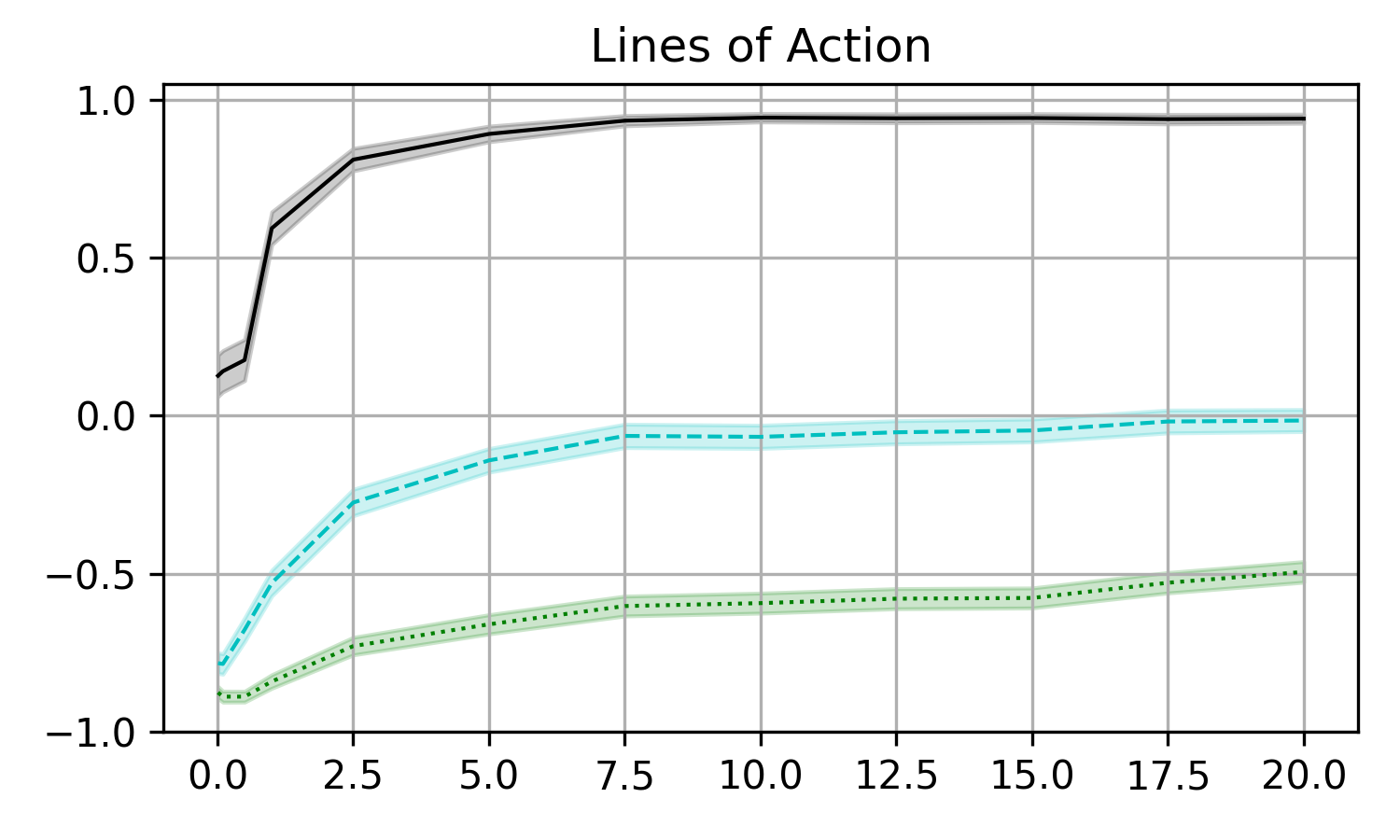}
\par\end{centering}
\begin{centering}
\includegraphics[scale=0.4]{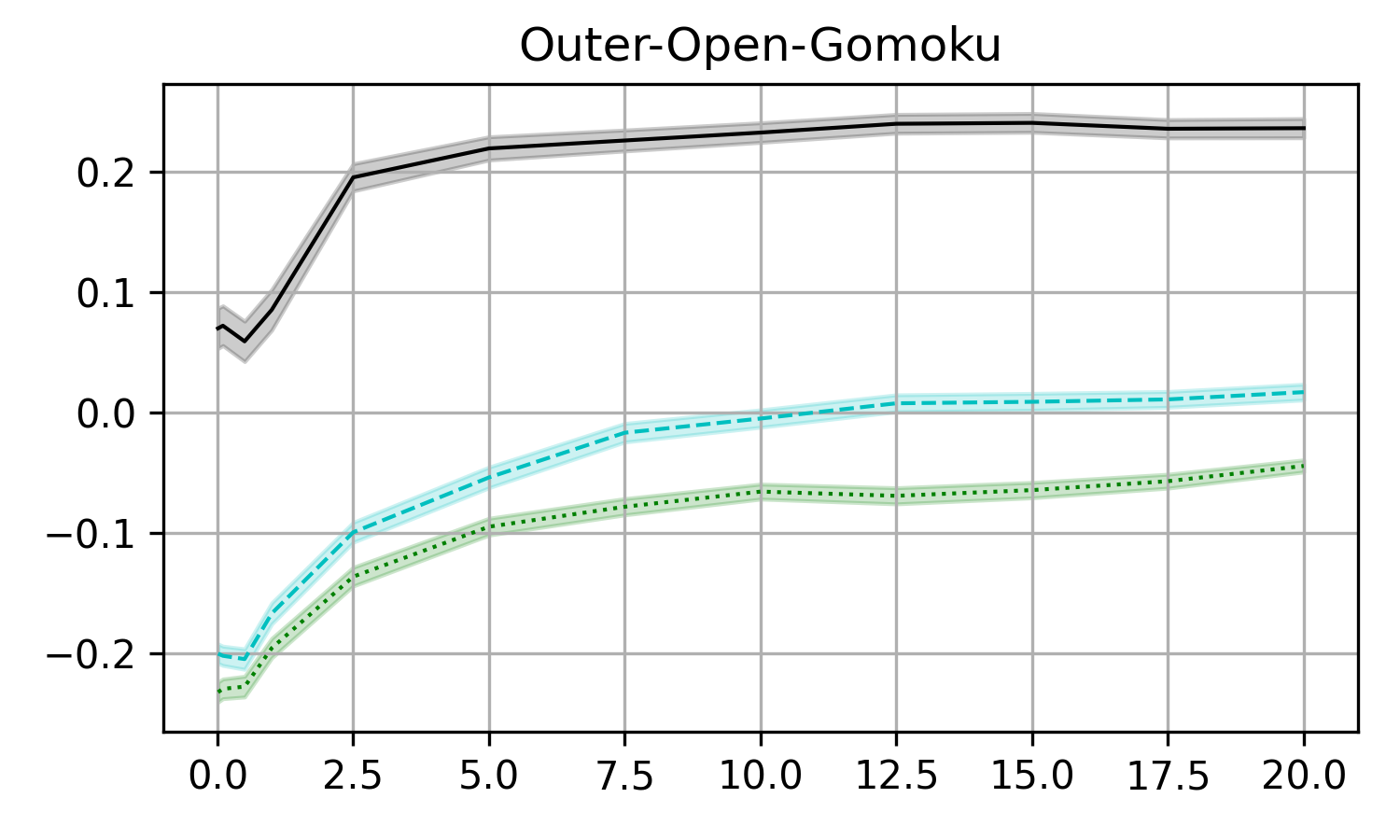}
\par\end{centering}
\begin{centering}
\includegraphics[scale=0.4]{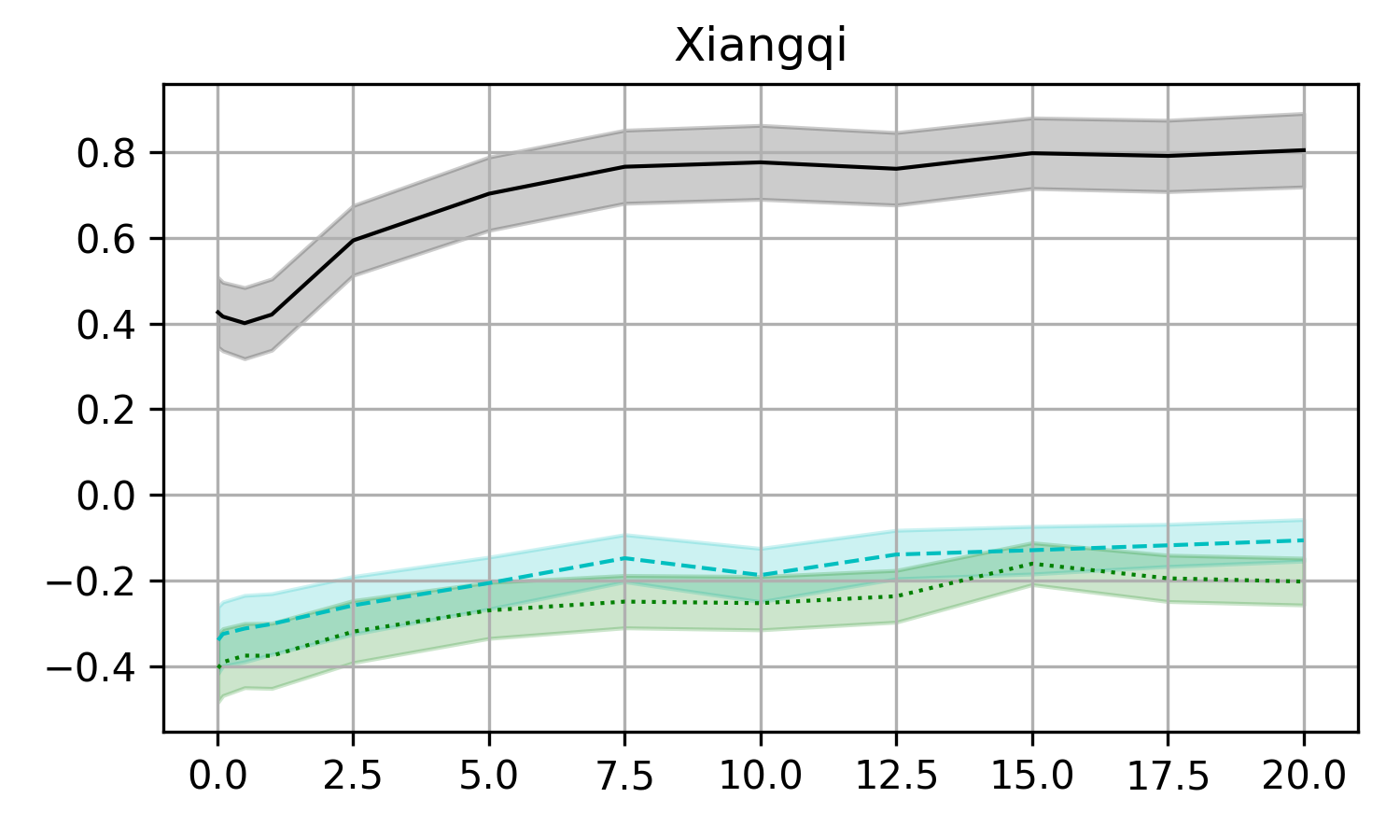}
\par\end{centering}
\caption{Average scores as a function of search time (in seconds) for \textcolor{teal}{$\protect\minibaln$
(dotted green curve)}, \textcolor{cyan}{$\protect\minibalp$ (dashed
cyan curve)}, and \textcolor{gray}{Unbounded Minimax (black line)}
across all studied games.}
\label{fig:courbe_moyenne-details-score}
\end{figure}

We now present the details of the results of the main paper experiment:
the evolution of the average win rate of the algorithms studied for
each game (Figure~\ref{fig:courbe_moyenne-details}) and the evolution
of their average scores for each game (Figure~\ref{fig:courbe_moyenne-details-score}).


\subsection{Additional Experience Against Very Weak Opponent}\label{subsec:Additional-Experience-Against}

We now present an additional experiment evaluating the performance
of the algorithms in the context of a very weak adversary: the standard
MCTS.

\subsubsection{Evaluation Protocol}

We evaluate Unbounded $\minibaln$, Unbounded $\minibalp$, and classical
Unbounded Minimax as a reference.

Each evaluated algorithm employs a high-level evaluation function
(as in the paper main experiment) and plays against the base MCTS
(using UCT and not using an evaluation function). Performance is averaged
over all high-level functions and over 20 repetitions.

Experiments are conducted across the same seven games. Overall, results
are averaged across all these games. 

\subsubsection{Technical Details}

For the balanced player, we evaluate multiple search budgets to study
how performance evolve as a function of this parameter.

We use the same strong 20 evaluation functions as for the main paper
experiment.

For each evaluated search time, an algorithm’s performance is measured
over 800 matches per game (20 high-level evaluation functions, considering
both first and second player positions, and repeted 20 times). 

The search times considered for the balanced algorithms were: $\left\{ 0.01,0.1,0.5,1,2.5,5,7.5,10,12.5,15,17.5,20\right\} $
seconds per move. The search time for MCTS is fixed at 1 second per
move. 

\subsubsection{Results}

We now present the performance of each algorithm according to the
binary performance metrics.

\begin{figure}
\begin{centering}
\includegraphics[scale=0.5]{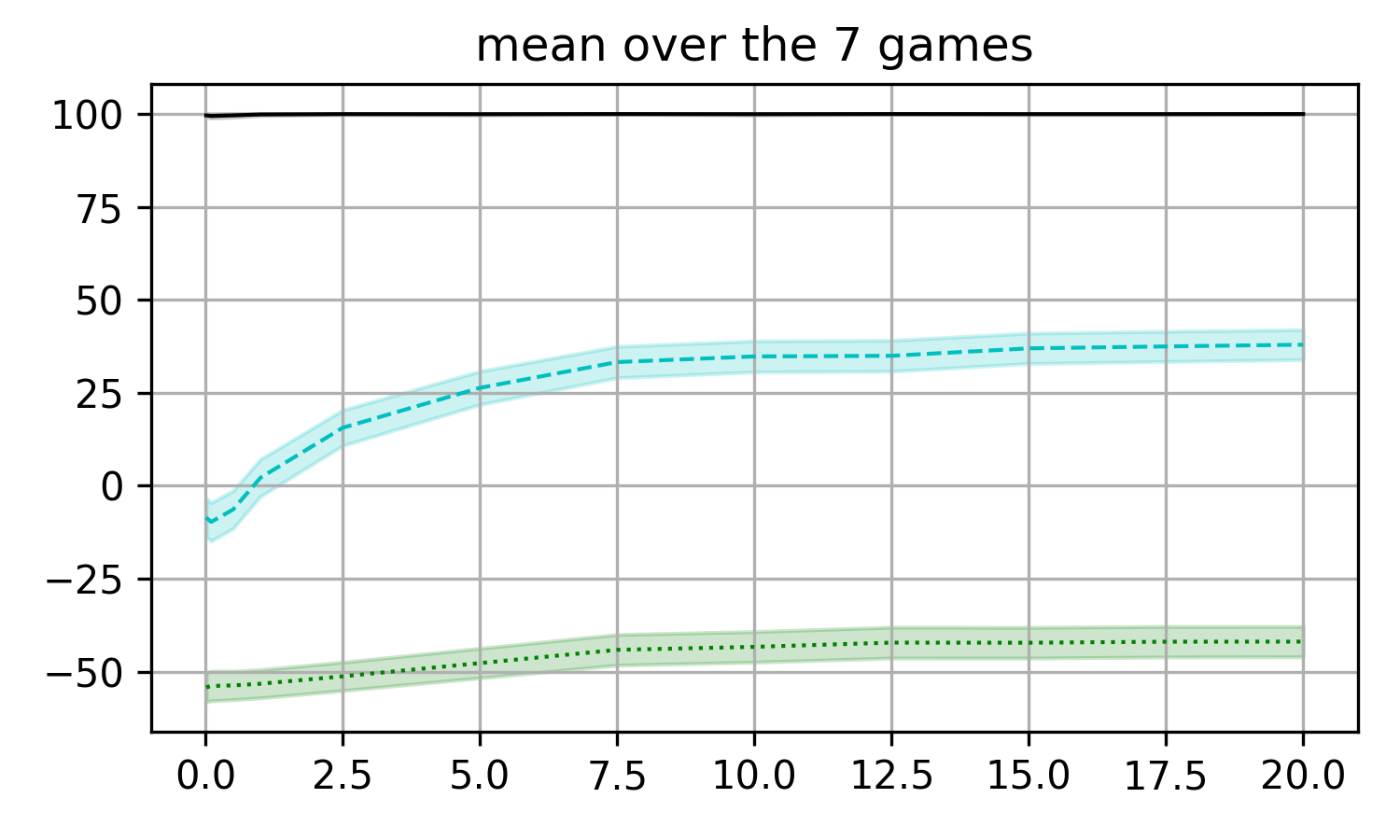}
\par\end{centering}
\caption{Average binary gain (left) and average scores (right) as a function
of search time (in seconds) for \textcolor{teal}{$\protect\minibaln$
(dotted green curve)}, \textcolor{cyan}{$\protect\minibalp$ (dashed
cyan curve)}, and \textcolor{gray}{Unbounded Minimax (black line)}
across the studied games against MCTS (in percentage).}\label{fig:Average-binary-gain-MCTS}

\end{figure}

The average binary gains (resp. average scores) are shown in Figure~\ref{fig:Average-binary-gain-MCTS},
which presents the outcomes as a function of search time per move.
Despite the very weak opponent, the algorithms manage to achieve a
significant balancing gain. The average binary gain for a search time
of 1 second per move are reported in Table~\ref{fig:Average-binary-gain-MCTS}.
For this parameter, the balancing is almost perfect for $\minibalp$.

\begin{table}
\centering{}{\tiny{}%
\begin{tabular}{|c|c|c|c|c|c|}
\hline 
{\tiny Unbounded} & {\tiny gain} & {\tiny 95\% C.R} & {\tiny win} & {\tiny draw} & {\tiny loss}\tabularnewline
\hline 
\hline 
{\tiny$\minibalp$} & {\tiny 2.16} & {\tiny 1.89} & {\tiny 27.16} & {\tiny 47.84} & {\tiny 25}\tabularnewline
\hline 
\end{tabular}}\caption{Average binary gains and win, draw, loss rates in percentage at 1
second of search time per move across the seven games.}
\label{tab:The-average-binary-mcts} 
\end{table}

\subsection{More Discussion}\label{subsec:More-Discussion}

We now provide additional discussions.

\subsubsection{Balanced Learning Problem}

Note that, in our experiments, the evaluation functions used for balancing
were trained to maximize winning probability. This choice raises an
important theoretical issue, even though our experiments proves highly
effective in practice.

On the one hand, a stronger player benefits from more accurate state-value
estimates, which improves its ability to control the game outcome
and thus to play in a balanced manner. On the other hand, as the skill
gap increases, the distribution of states encountered during matches
departs significantly from those seen during training, leading to
less reliable evaluations. This mismatch is particularly pronounced
when facing very weak opponents. In other words, the state space induced
by self-play training differs substantially from the state space encountered
when interacting with a much weaker player.

Consequently, learning to optimally master a game may partially conflict
with learning to play in a balanced way against a given target skill
level, especially when the state space is large and the target level
is far from optimal play.

This issue could be mitigated by developing evaluation functions explicitly
trained for balancing, which remains an open research problem, as
well as by addressing well-known challenges of neural and adaptive
function approximators, such as catastrophic forgetting and limited
generalization.

\subsubsection{Opponent of Equal Strength}

Note that if a balancing algorithm faces an algorithm that aims to
win, and both use the same evaluation function, the balancing algorithm
will have a lower effective playing strength. Indeed, on the one hand,
fewer resources are devoted to winning with the balancing algorithm,
since it seeks to play in a balanced manner; on the other hand, the
resulting strategy is closer to a draw and therefore more likely to
lead to a loss. It is thus preferable to use an evaluation function
that is stronger than the opponent’s, but not excessively so, as explained
in the previous section.
\end{document}